\documentclass{article}

\usepackage[final]{neurips_2020}

\usepackage[utf8]{inputenc} 
\usepackage[T1]{fontenc}    
\usepackage{hyperref}       
\usepackage{url}            
\usepackage{booktabs}       
\usepackage{amsfonts}       
\usepackage{amsmath}        
\usepackage{amssymb}
\usepackage{nicefrac}       
\usepackage{microtype}      
\usepackage{csquotes}       
\usepackage{subcaption}
\usepackage[pdftex]{graphicx} 
\usepackage{listings}
\usepackage{xcolor}
\usepackage{cleveref}
\usepackage{algorithm}
\usepackage[frozencache]{minted}

\newcommand\dd[2]{\frac{\partial #1}{\partial#2}}
\newcommand\Dd[2]{\frac{\mathrm{d} #1}{\mathrm{d} #2}}

\newcommand{\Ddn}[3]{\frac{\mathrm{d}^#3 #1}{\mathrm{d} #2^#3}}

\newcommand\jetcode[3]{\texttt{jet}(#1,{#2},({#3}))}
\newcommand\jettt{\texttt{jet}}
\newcommand\ff[1]{\frac{1}{#1!}}

\newcommand\R[1]{\mathbb{R}^{#1}}

\newcommand{\pfrac}[2]{\frac{\partial #1}{\partial #2}}
\newcommand{\pnfrac}[3]{\frac{\partial^{#3} #1}{\partial #2^{#3}}}
\newcommand{\ddfrac}[2]{\frac{\mathrm{d} #1}{\mathrm{d} #2}}
\newcommand{\dnfrac}[3]{\frac{d^{#3} #1}{d #2^{#3}}}

\newcommand{\factfrac}[1]{\frac{1}{#1!}}
\newcommand{\tayc}[1]{{[#1]}}
\newcommand\taycc[2]{#1_{\tayc{#2}}}
\newcommand\taycct[2]{\tilde{#1}_{\tayc{#2}}}

\newcommand\trans{\intercal}

\newcommand\mc[1]{\mathcal{#1}}
\newcommand\mb[1]{\mathbf{#1}}

\definecolor{vectorpink}{HTML}{E40086}
\definecolor{vectorlightblue}{HTML}{4C6379}
\definecolor{vectordarkblue}{HTML}{07223E}
\definecolor{niceblue}{HTML}{0074D9}
\definecolor{nicepink}{HTML}{B10DC9}
\definecolor{mydarkblue}{HTML}{001373}

\newcommand*{\titlemacro}{Learning differential equations that are easy to solve}

\hypersetup{
    pdftitle={Learning Differential Equations that are Easy to Solve},
    pdfauthor={},
    pdfsubject={Conference on Neural Information Processing Systems 2020},
    pdfkeywords={},
    pdfborder=0 0 0,
    pdfpagemode=UseNone,
    colorlinks=true,
    linkcolor=mydarkblue,
    citecolor=mydarkblue,
    filecolor=mydarkblue,
    urlcolor=mydarkblue,
    pdfview=FitH
}
\usepackage[title]{appendix}
\usepackage{wrapfig}

\usepackage{caption}
\title{\titlemacro}

\author{%
  Jacob Kelly\thanks{Equal Contribution.  Code available at:\newline \href{https://github.com/jacobjinkelly/easy-neural-ode}{\nolinkurl{ github.com/jacobjinkelly/easy-neural-ode}}\vspace{-2.1em}
} \\
  University of Toronto, Vector Institute\\
  \texttt{jkelly@cs.toronto.edu} \\
  \And
  Jesse Bettencourt\footnote[1]{Equal Contribution.} \\
  University of Toronto, Vector Institute\\
  \texttt{jessebett@cs.toronto.edu} \\
  \AND
  Matthew James Johnson \\
  \phantom{mmmmmm} Google Brain \phantom{mmmmmm}\\
  \texttt{mattjj@google.com} \\
  \And
  David Duvenaud \\
  University of Toronto, Vector Institute\\
  \texttt{duvenaud@cs.toronto.edu}
}

\begin{document}

\maketitle

\begin{abstract}
Differential equations parameterized by neural networks become expensive to solve numerically as training progresses.
We propose a remedy that encourages learned dynamics to be easier to solve. 
Specifically, we introduce a differentiable surrogate for the time cost of standard numerical solvers, using higher-order derivatives of solution trajectories.
These derivatives are efficient to compute with Taylor-mode automatic differentiation.
Optimizing this additional objective trades model performance against the time cost of solving the learned dynamics.
We demonstrate our approach by training substantially faster, while nearly as accurate, models in supervised classification, density estimation, and time-series modelling tasks.
\end{abstract}

\section{Introduction}

\begin{wrapfigure}[22]{r}{0.45\textwidth}  
    \centering
\vspace{-4em}
\begin{subfigure}[b]{0.99\linewidth}
  \centering
  \includegraphics[width=\linewidth]{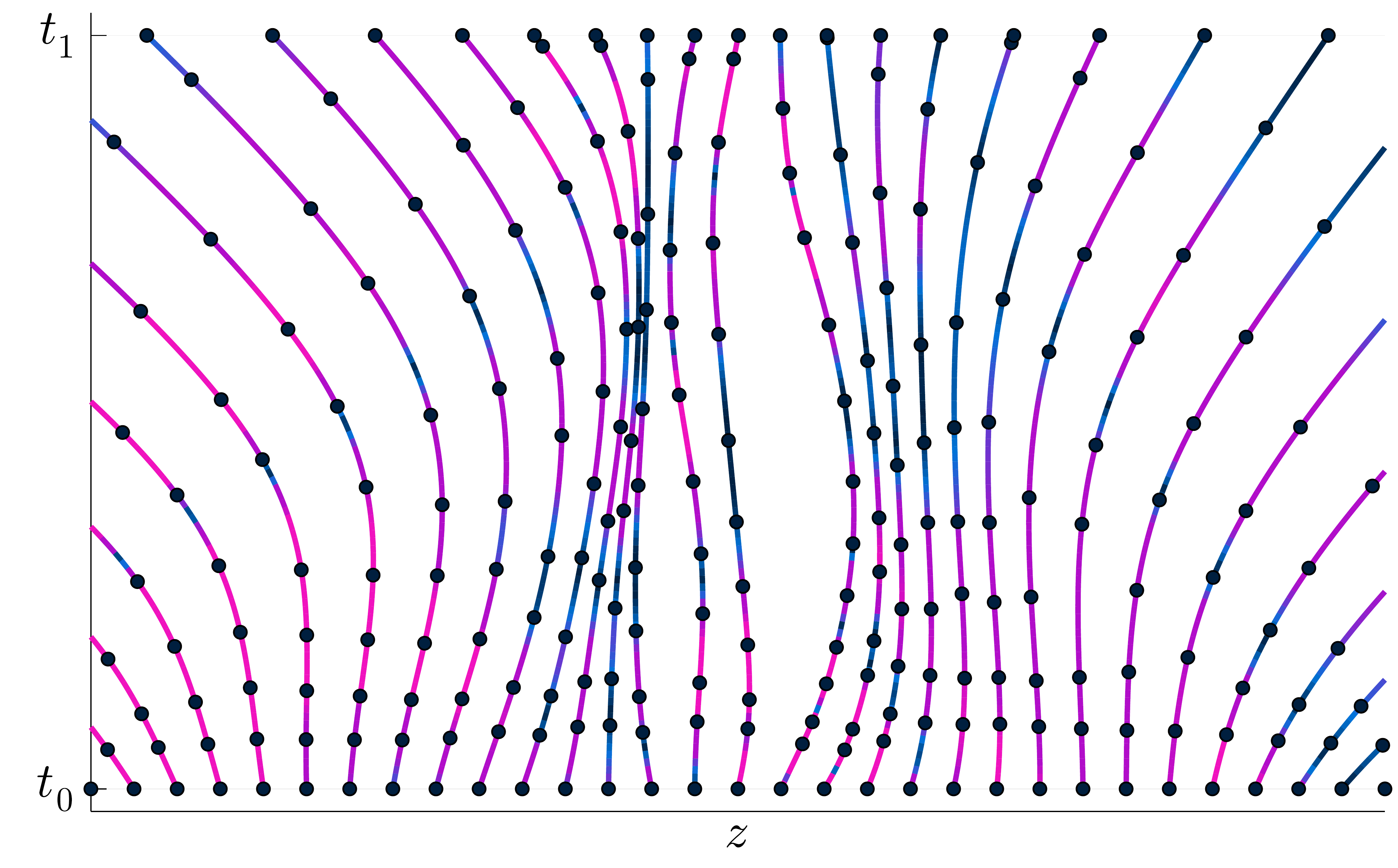}
    \end{subfigure}
    
\begin{subfigure}[b]{0.99\linewidth}
  \centering
  \includegraphics[width=\linewidth]{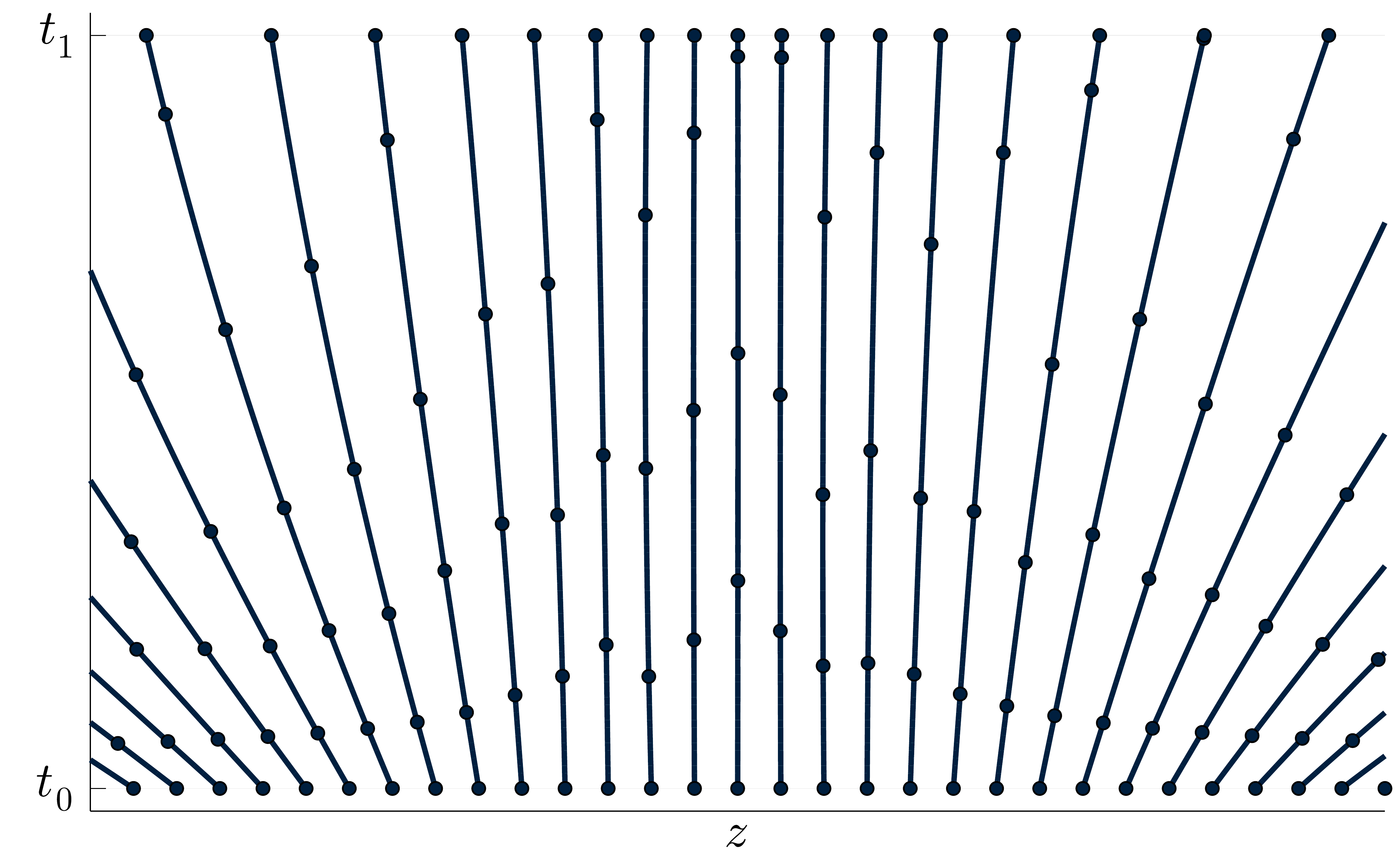}
    \end{subfigure}
    \vspace{-1.7em}
\caption{
\emph{Top:}
Trajectories of an ODE fit to map 
${\mb{z}(t_1) = \mb{z}(t_0) + \mb{z}(t_0)^3}$.
The learned dynamics are unnecessarily complex and require many evaluations (black dots) to solve. \\
\emph{Bottom:}
Regularizing the third total derivative $\Ddn{\mb{z}(t)}{t}{3}$ (shown by \textcolor{vectorpink}{colour})
gives dynamics that fit the same map, but require fewer evaluations to solve. 
}
\label{fig:jetregtoy}
\end{wrapfigure}


Differential equations describe a system's behavior by specifying its instantaneous dynamics.
Historically, differential equations have been derived from theory, such as Newtonian mechanics, Maxwell's equations, or epidemiological models of infectious disease, with parameters inferred from observations.
Solutions to these equations usually cannot be expressed in closed-form, requiring numerical approximation.

Recently, ordinary differential equations parameterized by millions of learned parameters, called neural ODEs, have been fit for latent time series models, density models, or as a replacement for very deep neural networks~\citep{rubanova2019latent, grathwohl2019ffjord, chen2018neural}.
These learned models are not constrained to match a theoretical model, only to optimize an objective on observed data. 
Learned models with nearly indistinguishable predictions can have substantially different dynamics.
This raises the possibility that we can find equivalent models that are easier and faster to solve.
Yet standard training methods have no way to 
penalize the complexity of the dynamics being learned.

How can we learn dynamics that are faster to solve numerically without substantially changing their predictions?
Much of the computational advantages of a continuous-time formulation come from using adaptive solvers, and most of the time cost of these solvers comes from repeatedly evaluating the dynamics function, which in our settings is a moderately-sized neural network.
So, we'd like to reduce the number of function evaluations (NFE) required for these solvers to reach a given error tolerance.
Ideally, we would add a term penalizing the NFE to the training objective, and let a gradient-based optimizer trade off between solver cost and predictive performance.
But because NFE is integer-valued, we need to find a differentiable surrogate. 

The NFE taken by an adaptive solver depends on how far it can extrapolate the trajectory forward without introducing too much error.
For example, for a standard adaptive-step Runge-Kutta solver with order $m$, the step size is approximately inversely proportional to the norm of the local $m$th total derivative of the solution trajectory with respect to time.
That is, a larger $m$th derivative leads to a smaller step size and thus more function evaluations.
Thus, we propose to minimize the norm of this total derivative during training, as a way to control the time required to solve the learned dynamics.

In this paper, we investigate the effect of this speed regularization in various models and solvers.
We examine the relationship between the solver order and the regularization order, and characterize the tradeoff between speed and performance.
In most instances, we find that solver speed can be approximately doubled without a substantial increase in training loss.
We also provide an extension to the JAX program transformation framework that provides Taylor-mode automatic differentiation, which is asymptotically more efficient for computing the required total derivatives than standard nested gradients.

Our work compares against and generalizes that of \citet{finlay2020train}, who proposed regularizing dynamics in the FFJORD density estimation model, and showed that it stabilized dynamics enough in that setting to allow the use of fixed-step solvers during training.

\section{Background}
\label{background}
An ordinary differential equation (ODE) specifies the instantaneous change of a vector-valued state $\mb{z}(t)$:
$\Dd{\mb{z}(t)}{t} = f(\mb{z}(t), t, \theta)$.
Given an initial condition $\mb{z}(t_0)$, computing the state at a later time:
$${\mb{z}(t_1) = \mb{z}(t_0) + \int_{t_0}^{t_1}f(\mb{z}(t),t,\theta) \, \mathrm{d} t}$$
is called an initial value problem (IVP).
For example, $f$ could describe the equations of motion for a particle, or the transmission and recovery rates for a virus across a population.
Usually, the required integral has no analytic solution, and must be approximated numerically.



\paragraph{Adaptive-step Runge-Kutta ODE Solvers}
Runge-Kutta methods~\citep{Runge, Kutta} approximate the solution trajectories of ODEs through a series of small steps, starting at time $t_0$.
At each step, they choose a step size $h$, and fit a local approximation to the solution, $\mb{\hat z}(t)$, using several evaluations of $f$.
When $h$ is sufficiently small, the numerical error of a $m$th-order method is bounded by
$\left\Vert\mb{\hat z}(t + h) - \mb{z}(t + h)\right\Vert \leq c h^{m+1}$
for some constant $c$~\citep{hairer}.
So, for a $m$th-order method, the local error grows approximately in proportion to the size of the $m$th coefficient in the Taylor expansion of the true solution.
All else being equal, controlling this coefficient for all dimensions of $\mb{z}(t)$ will allow larger steps to be taken without surpassing the error tolerance.

\paragraph{Neural Ordinary Differential Equations}
The dynamics function $f$ can be a moderately-sized neural network, and its parameters $\theta$ trained by gradient descent.
Solving the resulting IVP is analogous to evaluating a very deep residual network in which the number of layers corresponds to the number of function evaluations of the solver \citep{chang2017multi, ruthotto2018deep, chen2018neural}.
Solving such continuous-depth models using adaptive numerical solvers has several computational advantages over standard discrete-depth network architectures.
However, this approach is often slower than using a fixed-depth network, due to an inability to control the number of steps required by an adaptive-step solver.

\section{Regularizing Higher-Order Derivatives for Speed}
\label{method}
The ability of Runge-Kutta methods to take large and accurate steps is limited by the $K$th-order Taylor coefficients of the solution trajectory.
We would like these coefficients to be small.
Specifically, we propose to regularize the squared norm of the $K$th-order total derivatives of the state with respect to time, integrated along the entire solution trajectory:
\begin{align}
    \mathcal{R}_K(\theta)
    \label{eq:reg_integral}
    &= \int_{t_0}^{t_1} \left\Vert\Ddn{\mb{z}(t)}{t}{K}\right\Vert^2_2  \, \mathrm{d} t
\end{align}
where $\left\Vert \cdot \right\Vert_2^2$ is the squared $\ell_2$ norm, and the dependence on the dynamics parameters $\theta$ is implicit through the solution $\mb{z}(t)$ integrating $\ddfrac{\mb{z}(t)}{t} = f(\mb{z}(t),t,\theta)$.
During training, we weigh this regularization term by a hyperparameter $\lambda$ and add it to our original loss to get our regularized objective:
\begin{gather}
    L_{reg}(\theta) = L(\theta) + \lambda\mathcal{R}_K(\theta)
\end{gather}

\begin{wrapfigure}[19]{r}{0.35\linewidth}
\vspace{-1.5em}
  \centering
  \includegraphics[width=\linewidth, clip]{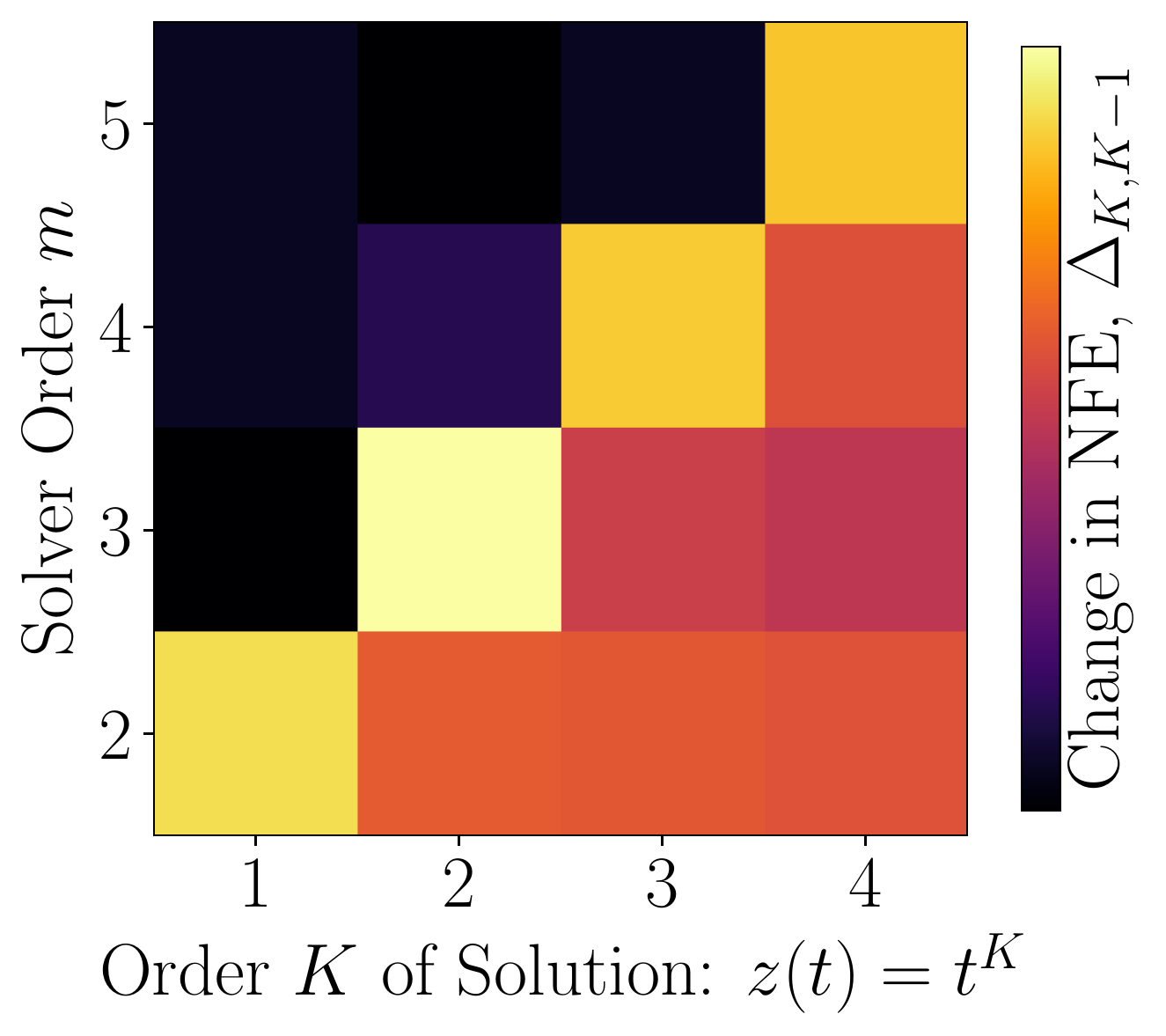}
\caption{$m$-order Runge-Kutta solvers need small steps when the dynamics have non-zero total derivatives of order $K\ge m$ (lower triangle).
Color denotes the increase in number of steps from $K$ to $K-1$, normalized  for each solver order.}
\label{fig:mnist_nfe_train1}
\end{wrapfigure}

What kind of solutions are allowed when $\mathcal{R}_K=0$?
For $K = 0$, we have $\left\Vert \mb{z}(t)\right\Vert_2^2 = 0$, so the only possible solution is $\mb{z}(t) = 0$.
For $K = 1$, we have $\left\Vert f(\mb{z}(t), t) \right\Vert_2^2 = 0$, so all solutions are constant, flat trajectories.
For $K = 2$ solutions are straight-line trajectories.
Higher values of $K$ shrink higher derivatives, but don't penalize lower-order dynamics.
For instance, a quadratic trajectory will have $\mathcal{R}_3 = 0$.
Setting the $K$th order dynamics to exactly zero everywhere automatically makes all higher orders zero as well.
\Cref{fig:jetregtoy} shows that regularizing $\mathcal{R}_3$ on a toy 1D neural ODE reduces NFE. 


Which orders should we regularize?
We propose matching the order of the regularizer to that of the solver being used.
We conjecture that regularizing dynamics of lower orders than that of the solver restricts the model unnecessarily, and that 
letting the lower orders remain unregularized should not increase NFE very much.
\Cref{fig:mnist_nfe_train1} shows empirically which orders of Runge-Kutta solvers can efficiently solve which orders of toy polynomial trajectories.
We partially confirm these conjectures on real models and datasets in \cref{order_reg}.

The solution trajectory and our regularization term can be computed in a single call to an ODE solver by augmenting the system with the integrand in \cref{eq:reg_integral}.

\section{Efficient Higher Order Differentiation with Taylor Mode}
The number of terms in higher-order forward derivatives grows exponentially in $K$, becoming prohibitively expensive for $K = 5$, and causing substantial slowdowns even for $K = 2$ and $K = 3$.
Luckily, there exists a generalization of forward-mode automatic differentiation (AD), known as Taylor mode, which can compute the total derivative exactly for a cost of only $\mathcal{O}(K^2)$. We found that this asymptotic improvement reduced wall-clock time by an order of magnitude, even for $K$ as low as 3.

\paragraph{First-order forward-mode AD} 
Standard forward-mode AD computes, for a function $f(x)$ and an input perturbation vector $v$,
the product $\pfrac{f}{x}v$.
This Jacobian-vector product, or JVP, can be computed efficiently without explicitly instantiating the Jacobian. 
This implicit computation of JVPs is straightforward whenever $f$ is a composition of operations for which which implicit JVP rules are known. 

\paragraph{Higher-order Jacobian-vector products}
Forward-mode AD can be generalized to higher orders to 
compute $K$th-order Jacobians contracted $K$ times against the perturbation vector: $\pnfrac{f}{x}{K} v^{\otimes K}$.
Similarly, this can also be computed without representing any Jacobian matrices explicitly. 

A na\" ive approach to higher-order forward mode is to recursively apply first-order forward mode.
Specifically, nesting JVPs $K$ times gives the right answer: ${\pnfrac{f}{x}{K} v^{\otimes K} = \pfrac{}{x}(\cdots (\pfrac{}{x}(\pfrac{f}{x}v)v)\cdots v)}$
but causes an unnecessary exponential slowdown, costing  $O(\exp(K))$.
This is because expressions that appear in lower derivatives also appear in higher derivatives, 
but the work to compute is not shared across orders.

\begin{wraptable}{r}{21em}
\vspace{-1em}
\centering
\begin{tabular}{c|c}
  Function & Taylor propagation rule \\
  \hline
  $y = z + cw$ & $\taycc{y}{k} = \taycc{z}{k} + c \taycc{w}{k}$\\
  $y = z * w$ & $\taycc{y}{k} = \sum_{j=0}^k \taycc{z}{j}\taycc{w}{k-j}$\\
  $y = z / w$ & $\taycc{y}{k} = \frac{1}{w_0} \left[z_k - \sum_{j=0}^{k-1} \taycc{z}{j}\taycc{w}{k-j}\right]$\\
	$y = \exp(z)$ & $\taycct{y}{k} = \sum_{j=1}^k \taycc{y}{k-j}\taycct{z}{j}$\\
	$s = \sin(z)$ & $\taycct{s}{k} = \sum_{j=1}^k \taycct{z}{j}\taycc{c}{k-j}$\\
	$c = \cos(z)$ & $\taycct{c}{k} = \sum_{j=1}^k -\taycct{z}{j}\taycc{s}{k-j}$\\
\end{tabular}
\caption{
Rules for propagating Taylor polynomial coefficients through standard functions.
These rules generalize standard first-order derivatives.
Notation $\taycc{z}{i} = \ff{i} z_i$
and $\taycct{y}{i} = \frac{i}{i!} z_i$. 
} \vspace{-2em}
\label{tab:prims}
\end{wraptable}

\paragraph{Taylor Mode}
Taylor-mode AD generalizes first-order forward mode to compute the first $K$ derivatives exactly with a time cost of only $O(K^2)$ or $O(K \log K)$, depending on the operations involved.
Instead of providing rules for propagating perturbation vectors,
one provides rules for propagating truncated Taylor series.
Some example rules are shown in \cref{tab:prims}.
For more details see the Appendix
and \citet[Chapter 13]{evalderivs}.
We provide an open source implementation of Taylor mode AD in the JAX Python library \citep{jax2018github}.

\section{Experiments}
\label{experiments}

\begin{wrapfigure}[15]{r}{0.4\linewidth}
\vspace{-3em}
  \centering
  \includegraphics[width=\linewidth]{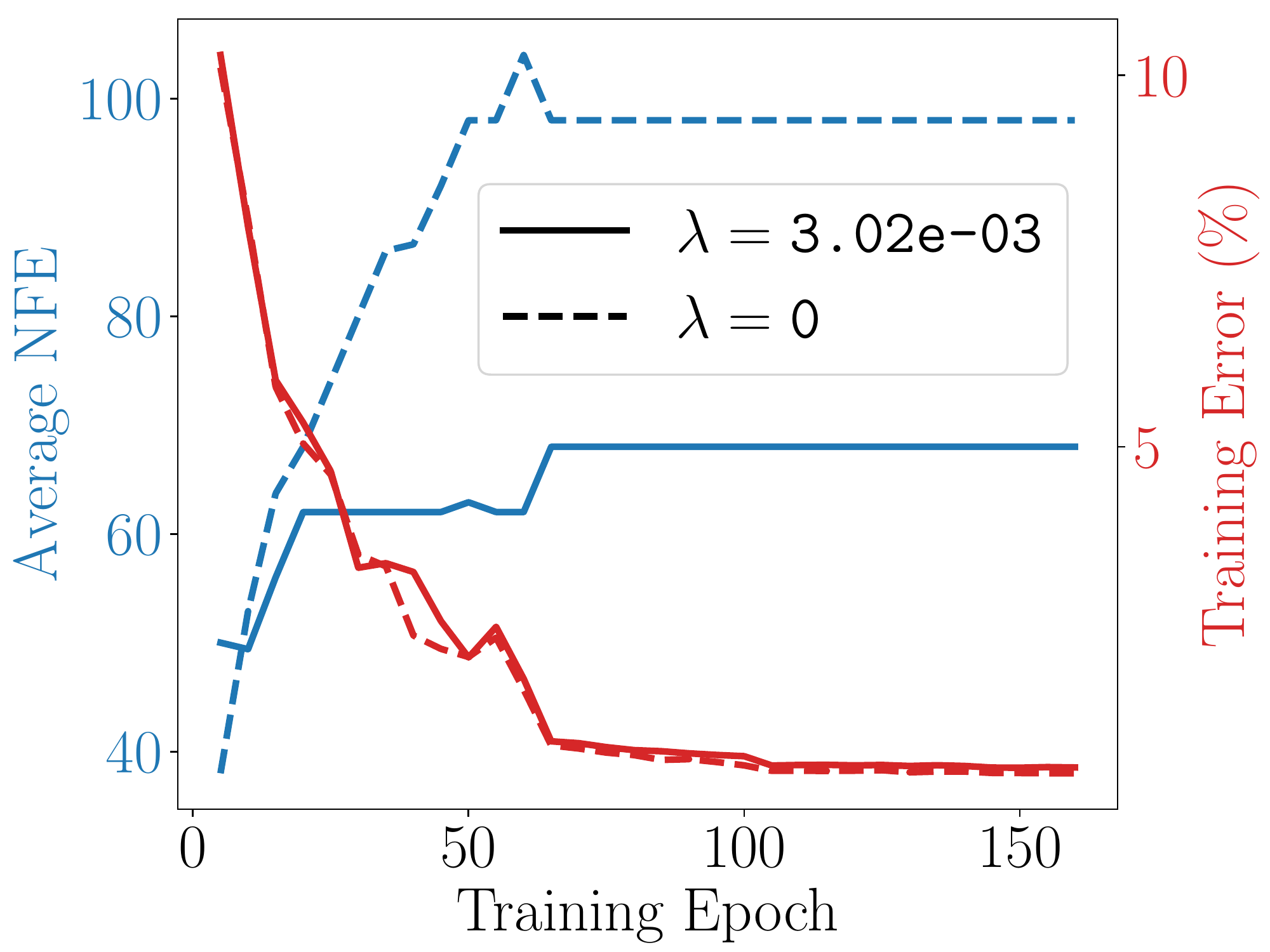}
\caption{Number of function evaluations (NFE) and training error during training.
Speed regularization (solid) decreases the NFE throughout training without substantially changing the training error.}
\label{fig:mnist_nfe_train2}
\end{wrapfigure}

We consider three different tasks in which continuous-depth or continuous time models might have computational advantages over standard discrete-depth models: supervised learning, continuous generative modeling of time-series \citep{rubanova2019latent}, and density estimation using continuous normalizing flows \citep{grathwohl2019ffjord}.
Unless specified otherwise, we use the standard \texttt{dopri5} Runge-Kutta 4(5) solver~\citep{dormand1980family, rk45_shampine}.

\subsection{Supervised Learning}
\label{sec:suplearn}
We construct a model for MNIST classification: it takes in as input a flattened MNIST image and integrates it through dynamics given by a simple MLP, then applies a linear classification layer.
In \cref{fig:mnist_nfe_train2} we compare the NFE and training error of a model with and without regularizing $\mc{R}_3$.

\begin{wrapfigure}[14]{r}{0.45\linewidth}
\vspace{-4.5em}
  \centering
  \begin{subfigure}[c]{0.49\linewidth}
  \centering
  \includegraphics[height=1\linewidth, clip, trim=1cm 0.1cm 1cm 0.1cm]{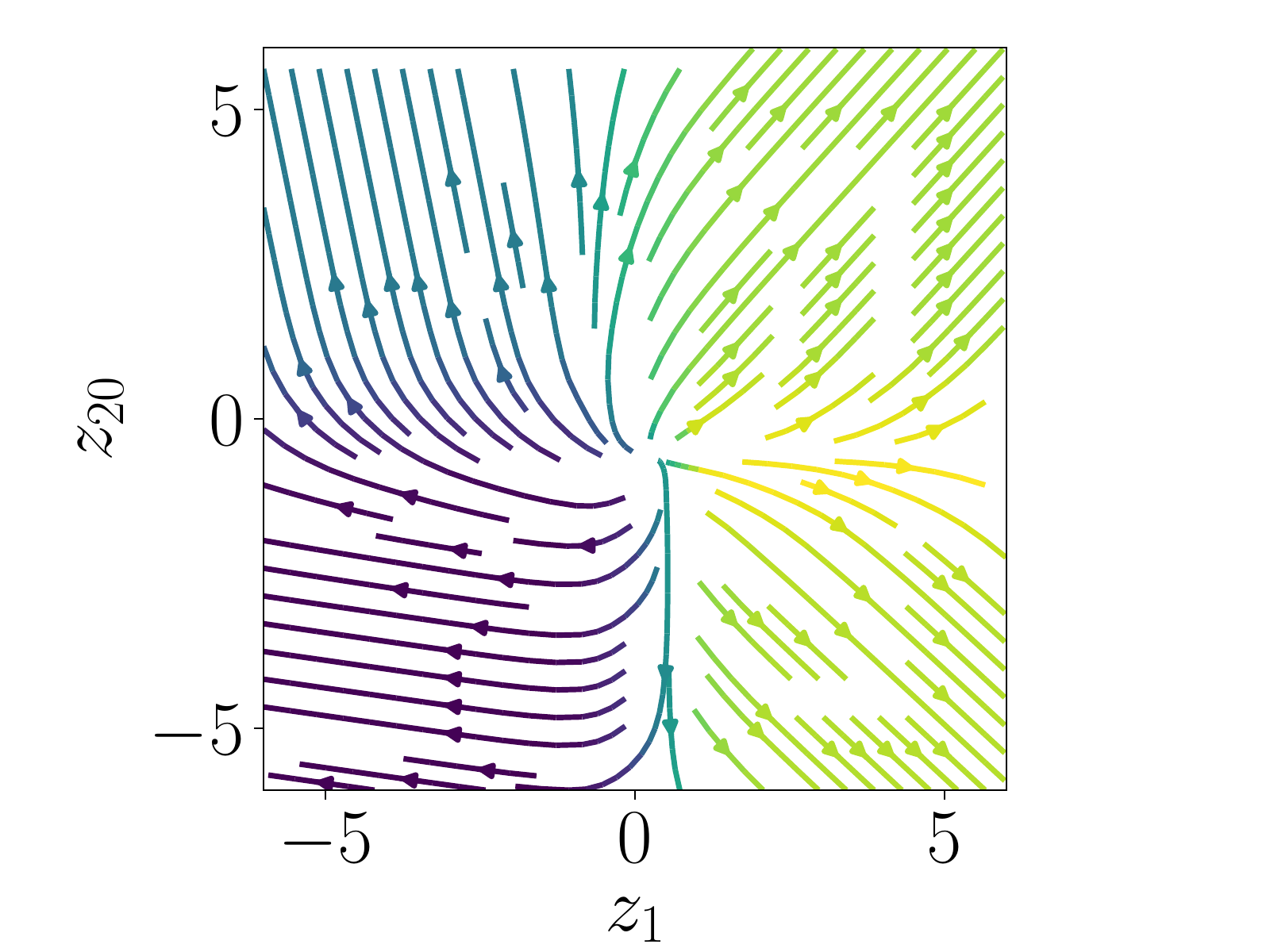}
  \caption{Unregularized}
  \end{subfigure}
  \begin{subfigure}[c]{0.49\linewidth}
  \centering
  \includegraphics[height=1\linewidth, clip, trim=3.3cm 0.1cm 1cm 0.1cm]{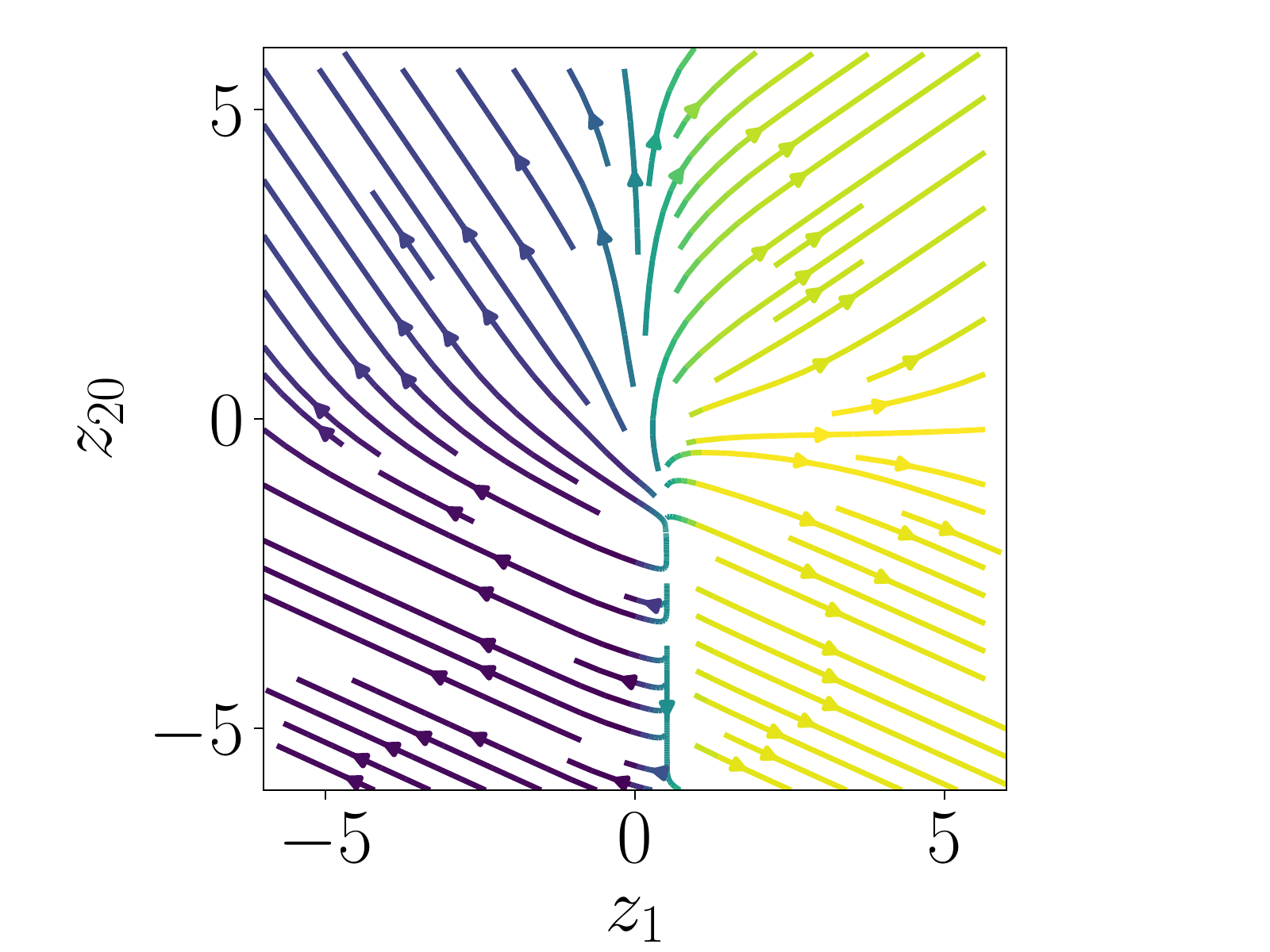}
   \caption{Regularized}
  \end{subfigure}
\caption{Regularizing dynamics in a latent ODE modeling PhysioNet clinical data.
Shown are a representative 2-dimensional slice of 20 dimensional dynamics.
We reduce average NFE from 281 to 90 while only incurring an 8\% increase in loss.} \label{fig:latent_dyn}
\end{wrapfigure}

\subsection{Continuous Generative Time Series Models}
As in \citet{rubanova2019latent}, we use the Latent ODE architecture for modelling trajectories of ICU patients using the PhysioNet Challenge 2012 dataset~\citep{physionet}.
This variational autoencoder architecture uses an RNN recognition network, and models the state dynamics using an ODE in a latent space.

In the supervised learning setting described in the previous section only the final state affects model predictions.
In contrast, time-series models' predictions also depend on the value of the trajectory at all intermediate times when observations were made.
So, we might expect speed regularization to be ineffective due to these extra constraints on the dynamics.
However, \cref{fig:latent_dyn} shows that, without changing their overall shape the latent dynamics can be adjusted to reduce their NFE by a factor of 3.

\subsection{Density Estimation with Continuous Normalizing Flows}
Our third task is unsupervised density estimation, using a scalable variant of continuous normalizing flows called FFJORD~\citep{grathwohl2019ffjord}.
We fit the MINIBOONE tabular dataset from \citet{tabularpapamakarios} and the MNIST image dataset~\citep{lecun2010mnist}.
We use the respective singe-flow architectures from \citet{grathwohl2019ffjord}.

\citet{grathwohl2019ffjord} noted that the NFE required to numerically integrate their dynamics could become prohibitively expensive throughout training.
\Cref{table:ffjord_mnist} shows that we can reduce NFE by 38\% for only a 0.6\% increase in log-likelihood measured in bits/dim. 

\paragraph{How to train your Neural ODE}
We compare against the approach of \citet{finlay2020train}, who design two regularization terms specifically for stabilizing the dynamics of FFJORD models:
\begin{align}
    \mathcal{K}(\theta) &= \int_{t_0}^{t_1} \left\Vert f(\mb{z}(t), t, \theta) \right\Vert^2_2 \, \mathrm{d}t \label{fin_kin} \\
    \mathcal{B}(\theta) &= \int_{t_0}^{t_1} \left\Vert \epsilon^\trans \nabla_\mb{z} f(\mb{z}(t), t, \theta) \right\Vert^2_2 \, \mathrm{d}t , \qquad \epsilon \sim \mathcal{N}(0, I) \label{fin_fro}
\end{align}
The first term is designed to encourage straight-line paths, and the second, stochastic, term is designed to reduce overfitting.
\citet{finlay2020train} used fixed-step solvers during training for some datasets.
We compare these two regularization on training with each of adaptive and fixed-step solvers, and evaluated using an adaptive solver, in \cref{finlay_comp}.

\section{Analysis and Discussion}

\subsection{Trading off function evaluations for loss}

\begin{figure}[ht!]
\vspace{-1em}
    \centering
    \begin{subfigure}[b]{0.32\linewidth}
        \centering
        \includegraphics[width=1\linewidth, 
        clip, trim=0cm 0.1cm 1.15cm 0.4cm]{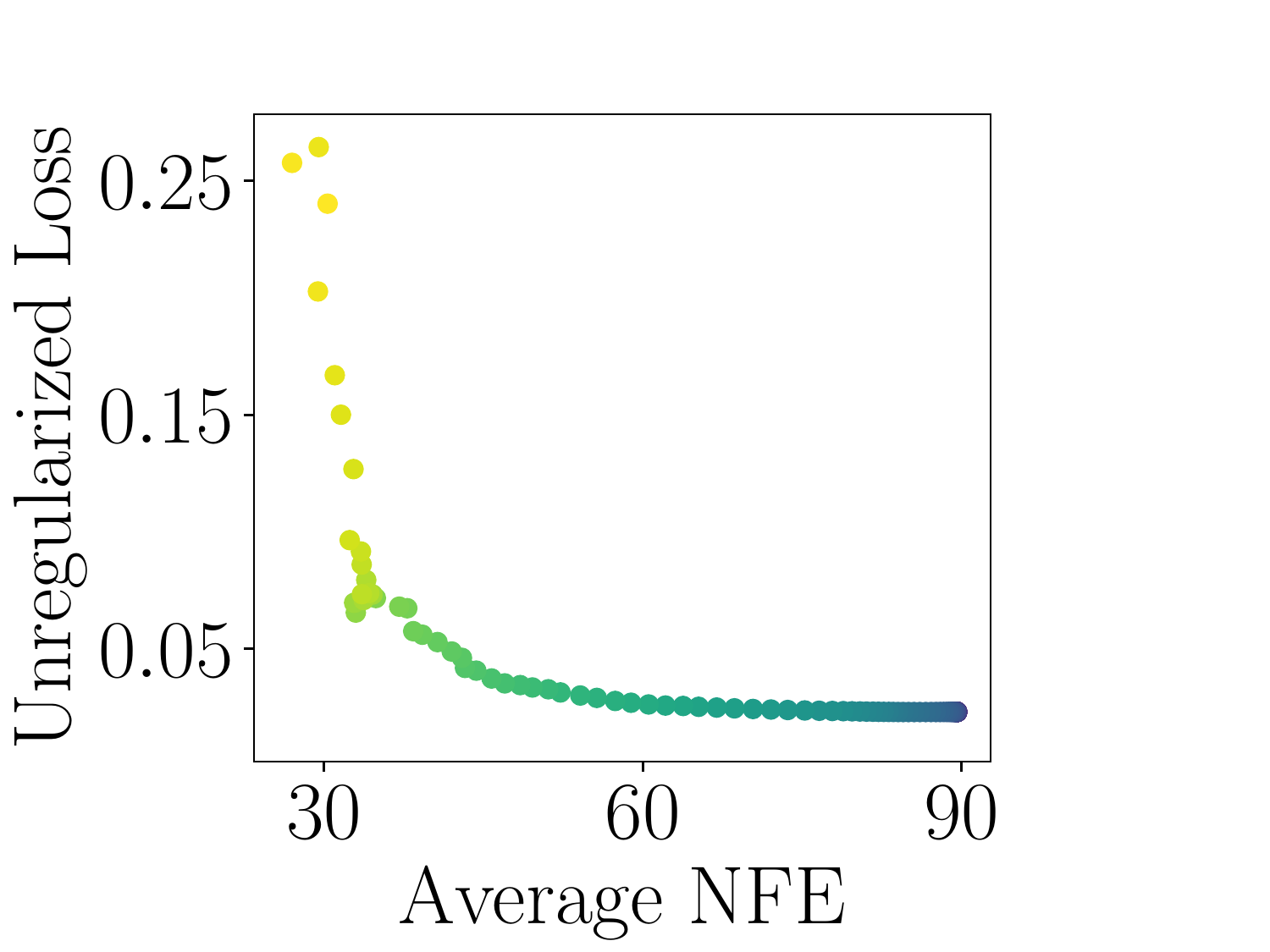}
        \caption{MNIST Classification}
    \end{subfigure}
    \begin{subfigure}[b]{0.32\linewidth}
        \centering
        \includegraphics[width=1\linewidth, 
        clip, trim=0.575cm 0.1cm 0.575cm 0.4cm]{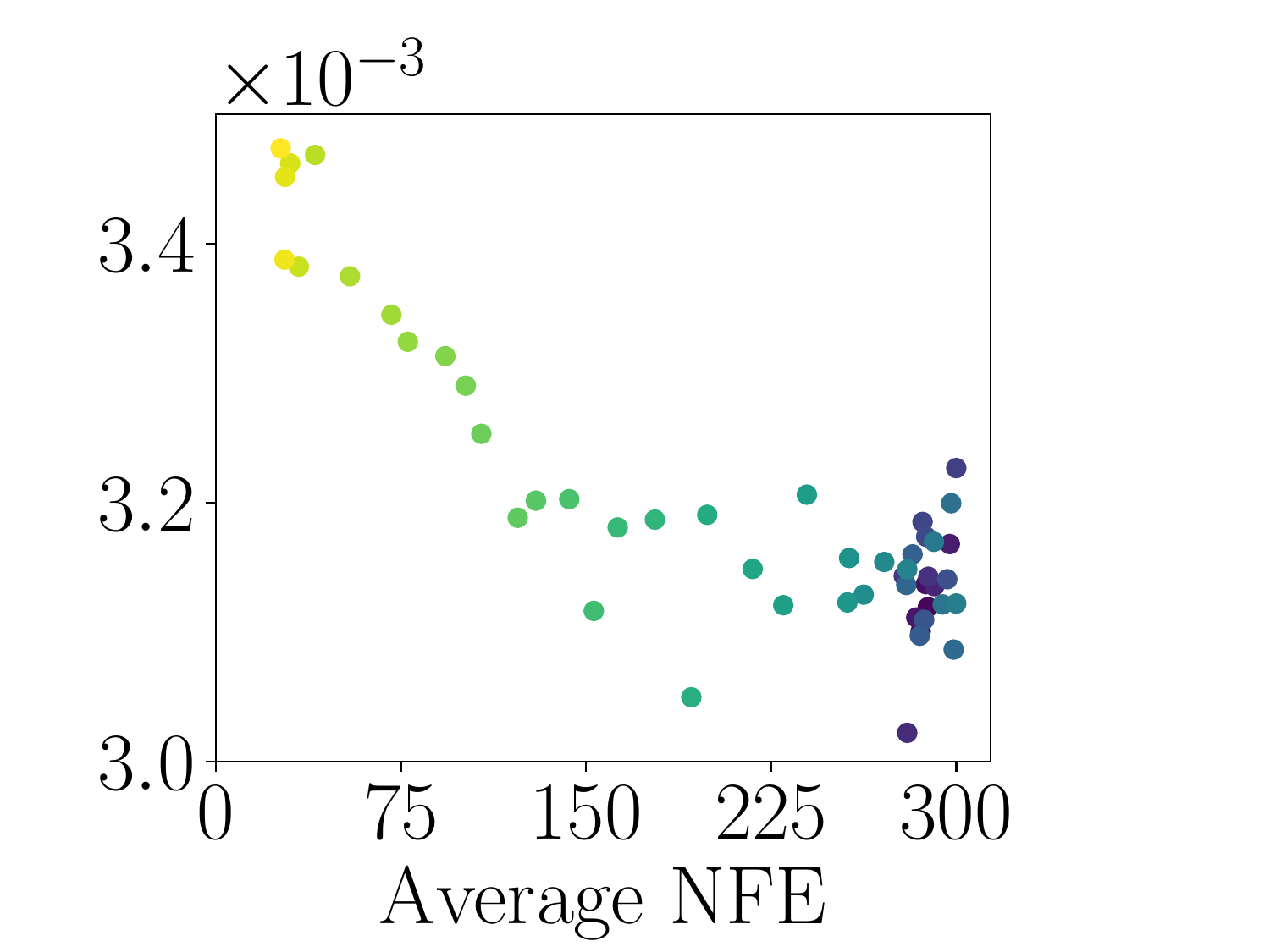}
        \caption{PhysioNet Time-Series}
    \end{subfigure}
    \begin{subfigure}[b]{0.32\linewidth}
        \centering
        \includegraphics[width=1\linewidth, 
        clip, trim=1.15cm 0cm 0.1cm 0.4cm
	   ]{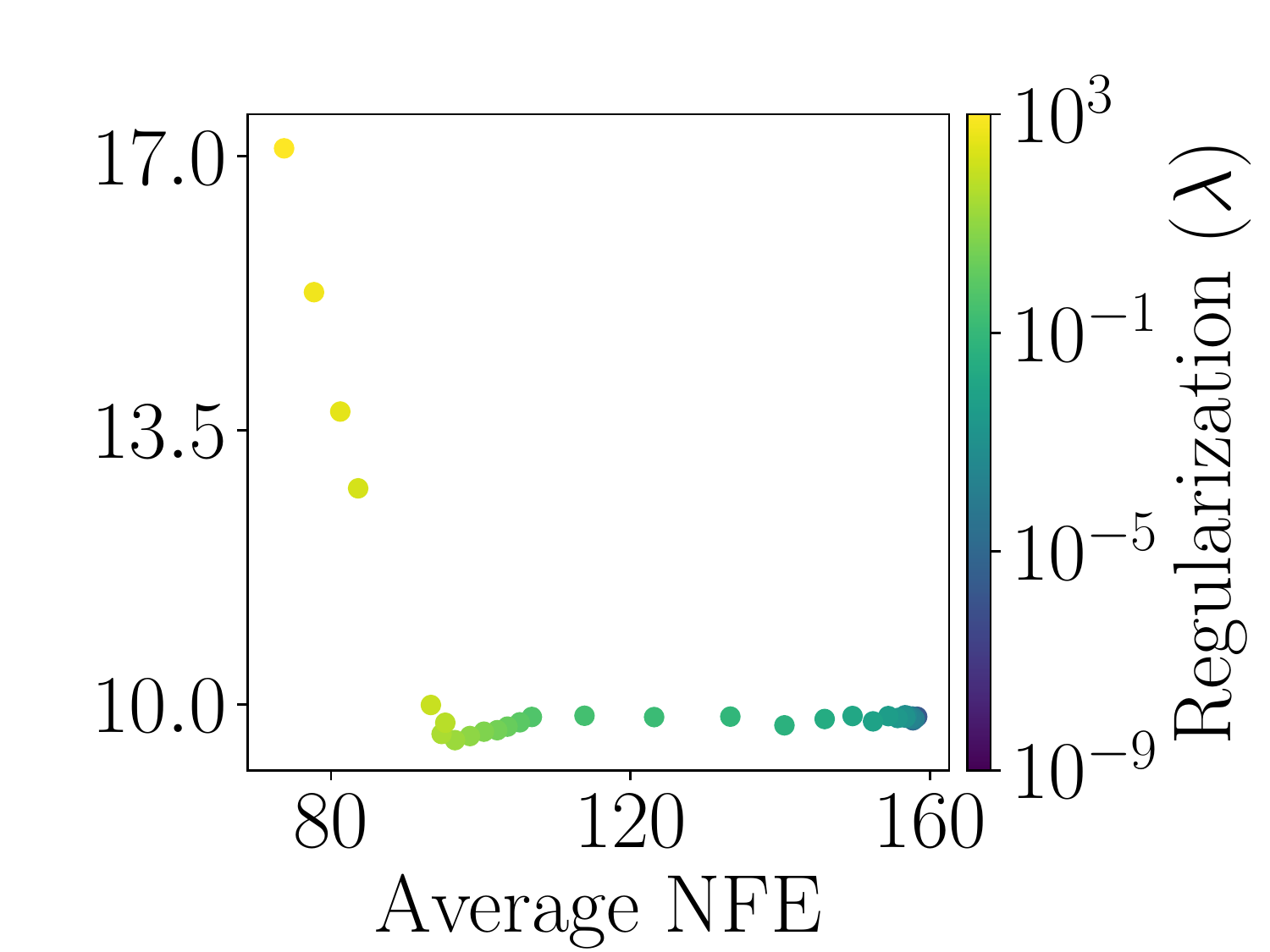}
        \caption{Miniboone Density Estimation}
    \end{subfigure}
\captionsetup{belowskip=-17pt}
\caption{Tuning the regularization of $\mathcal{R}_2$ trades off between training loss and solver speed in three different applications of neural ODEs.
Horizontal axes show average number of function evaluations, and vertical axes show unregularized training loss, both at the end of training.
}
\label{fig:paretos}
\end{figure}

What does the trade off between accuracy and speed look like?
Ideally, we could reduce the solver time a lot without substantially reducing model performance.
Indeed, this is demonstrated in all three settings we explored.
\Cref{fig:paretos} shows that generally, model performance starts getting substantially worse only after a 50\% reduction in solver speed when controlling $\mathcal{R}_2$.

\subsection{Order of regularization vs. order of solver}
\label{order_reg}
Which order of total derivatives should we regularize for a particular solver?
As mentioned earlier, we conjecture that the best choice would be to match the order of the solver being used.
Regularizing too low an order might needlessly constrain the dynamics and make it harder to fit the data, while regularizing too high an order might leave the dynamics difficult to solve for a lower-order solver.
However, we also expect that optimizing higher-order derivatives might be challenging, since these higher derivatives can change quickly even for small changes to the dynamics parameters.

\begin{figure}[ht]
\vspace{-2em}
    \centering
    \begin{subfigure}[b]{0.24\linewidth}
        \centering
        \includegraphics[width=1\linewidth, 
        clip, trim=0.17cm 0.15cm 1.8cm 1.25cm]{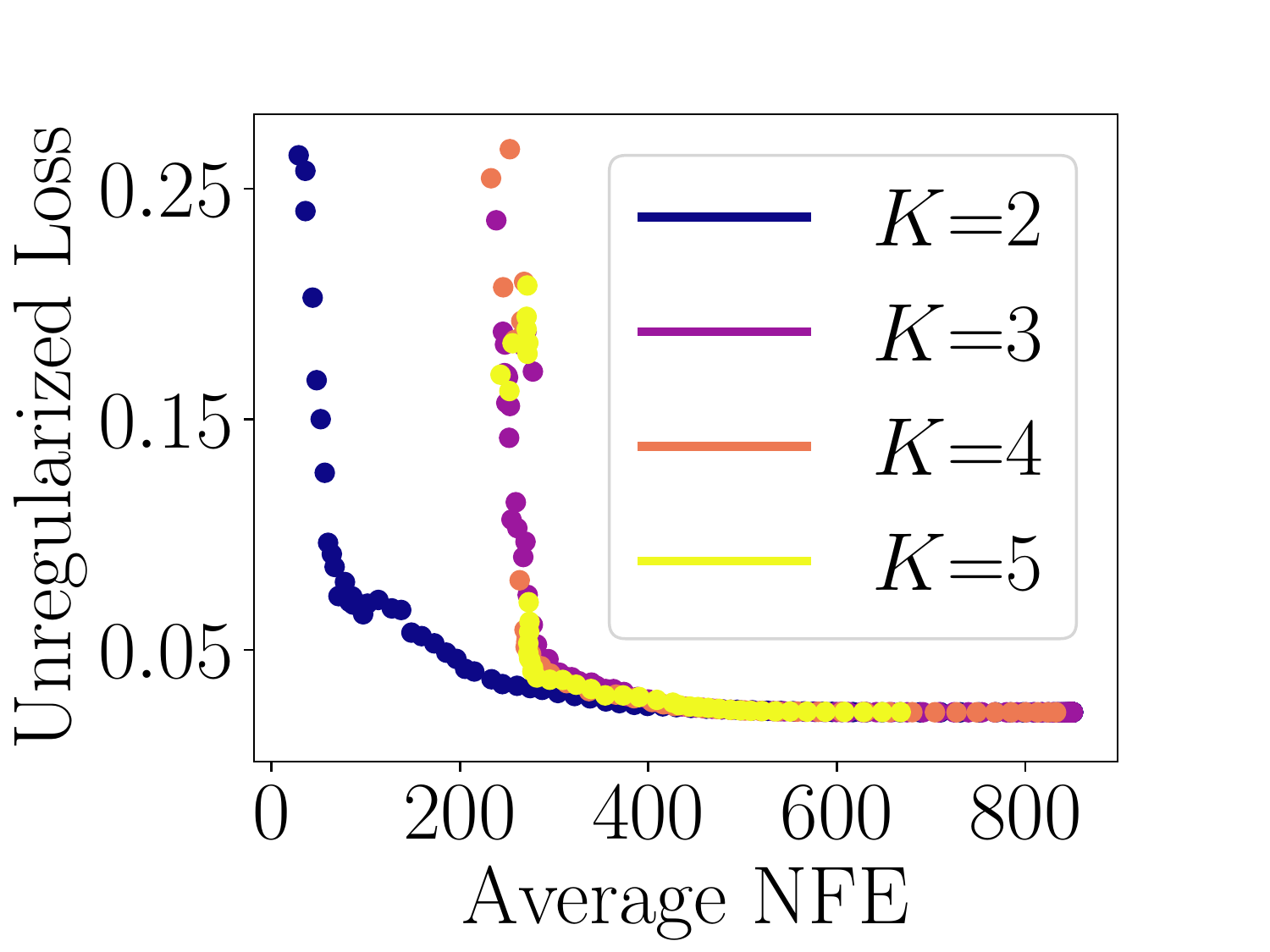}
        \caption{Order 2 Solver}
        \label{fig:orders1}
    \end{subfigure}
    \begin{subfigure}[b]{0.24\linewidth}
        \centering
        \includegraphics[width=1\linewidth, 
        clip, trim=0.17cm 0.15cm 1.8cm 1.25cm]{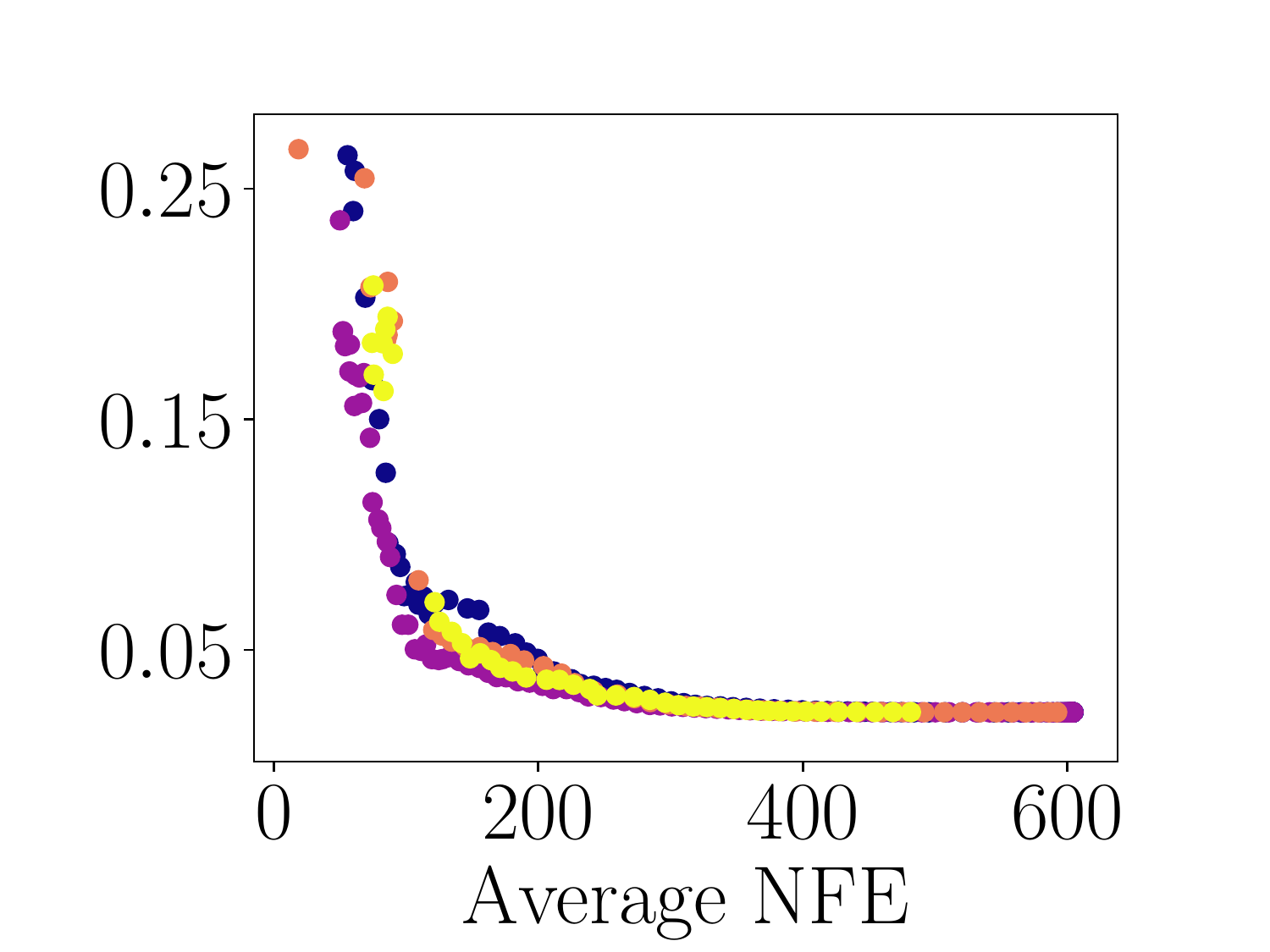}
        \caption{Order 3 Solver}
        \label{fig:orders2}
    \end{subfigure}
    \begin{subfigure}[b]{0.24\linewidth}
        \centering
        \includegraphics[width=1\linewidth, 
        clip, trim=0.17cm 0.15cm 1.8cm 1.25cm]{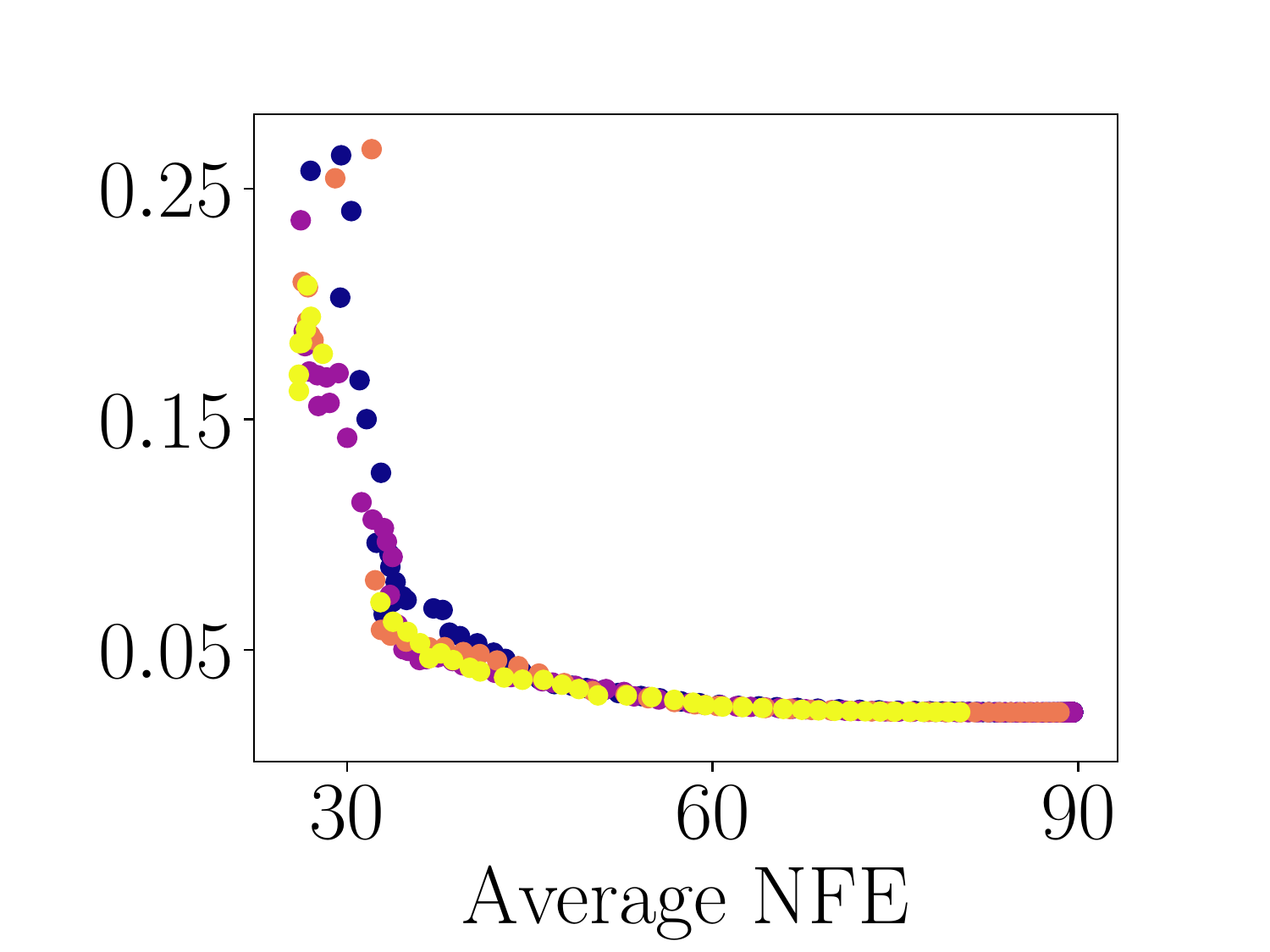}
        \caption{Order 5 Solver}
        \label{fig:orders3}
    \end{subfigure}
    \begin{subfigure}[b]{0.24\linewidth}
        \centering
        \includegraphics[width=1\linewidth, 
        clip, trim=0.17cm 0.15cm 1.8cm 1.25cm]{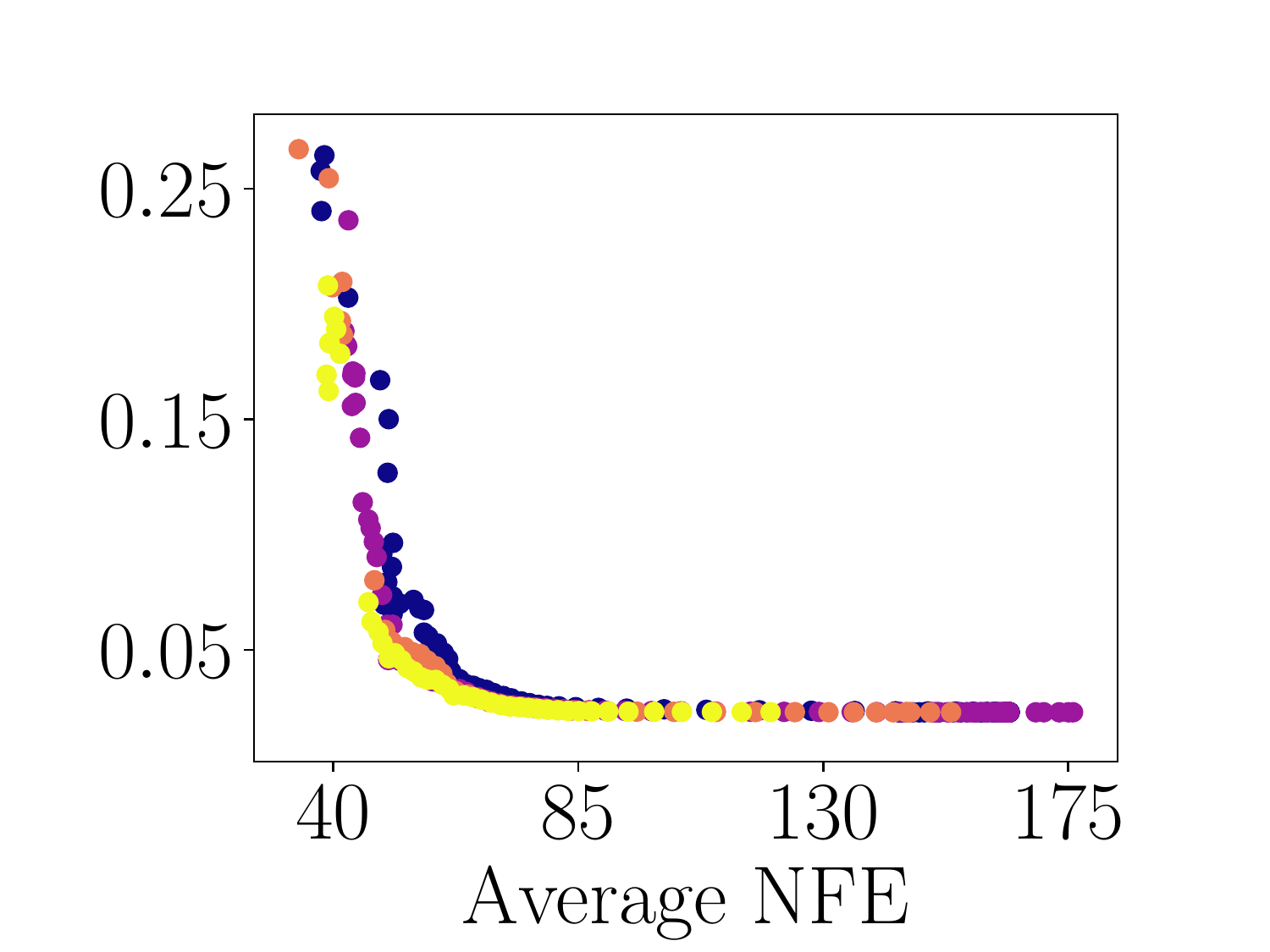}
        \caption{Adaptive Order Solver}
        \label{fig:orders4}
    \end{subfigure}
\caption{Comparing tradeoff between speed and performance when regularizing different orders.
\ref{fig:orders1}): For a 2nd-order solver, regularizing the 2nd total derivative gives the best tradeoff.
\ref{fig:orders2}): For a 3rd-order solver, regularizing the 3rd total derivative gives the best tradeoff, but the difference is small.
\ref{fig:orders3}): For a 5th-order solver, results are mixed.
\ref{fig:orders4}): For an adaptive-order solver, the difference is again small but regularizing higher orders works slightly better.
}
\label{fig:orders}
\end{figure}

\begin{figure}[t]
\vspace{-.5em}
    \centering
    \begin{subfigure}[b]{0.32\linewidth}
        \centering
        \includegraphics[width=1\linewidth]{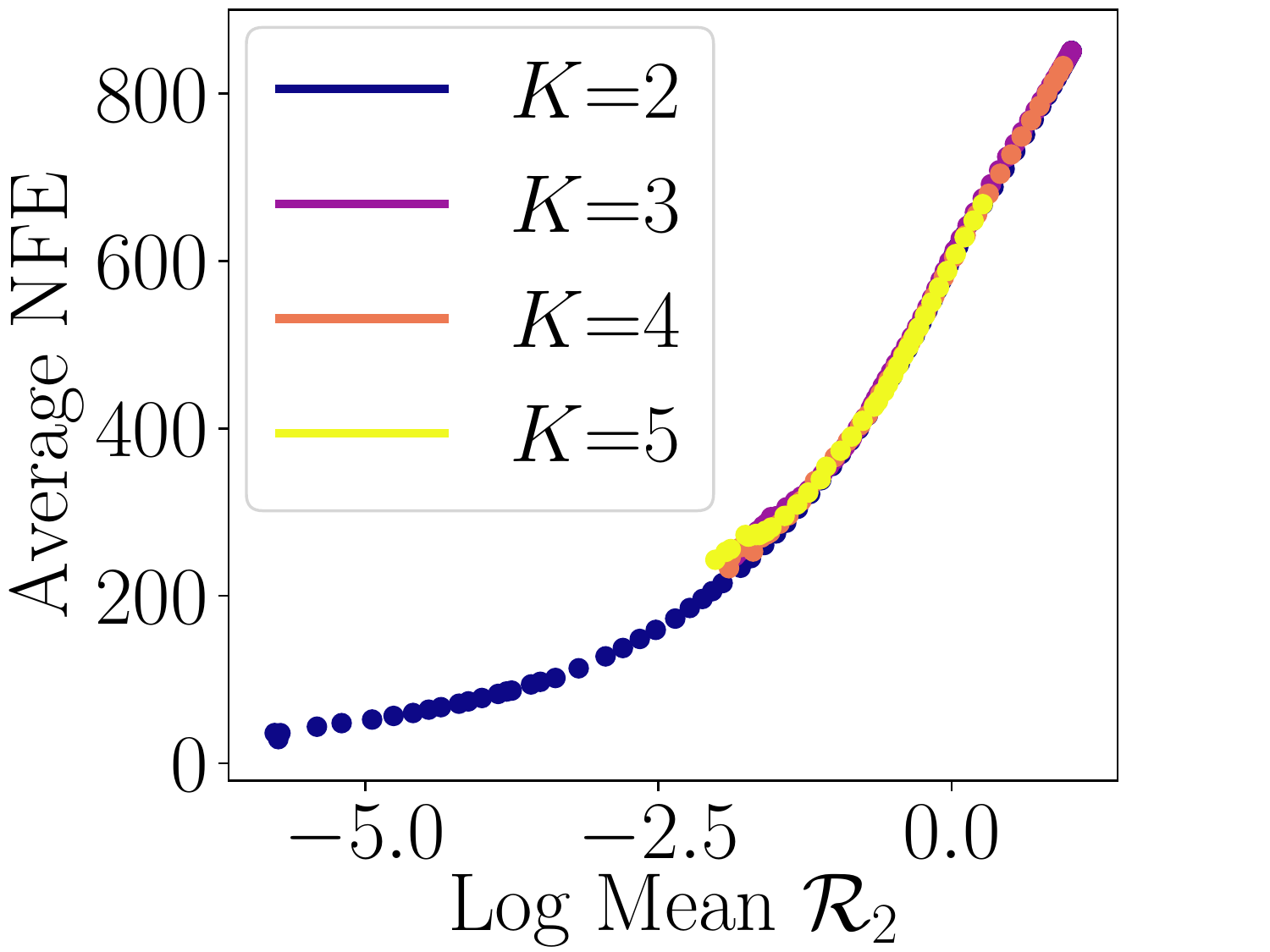}
        \caption{Order 2 Solver}
    \end{subfigure}
    \begin{subfigure}[b]{0.32\linewidth}
        \centering
        \includegraphics[width=1\linewidth]{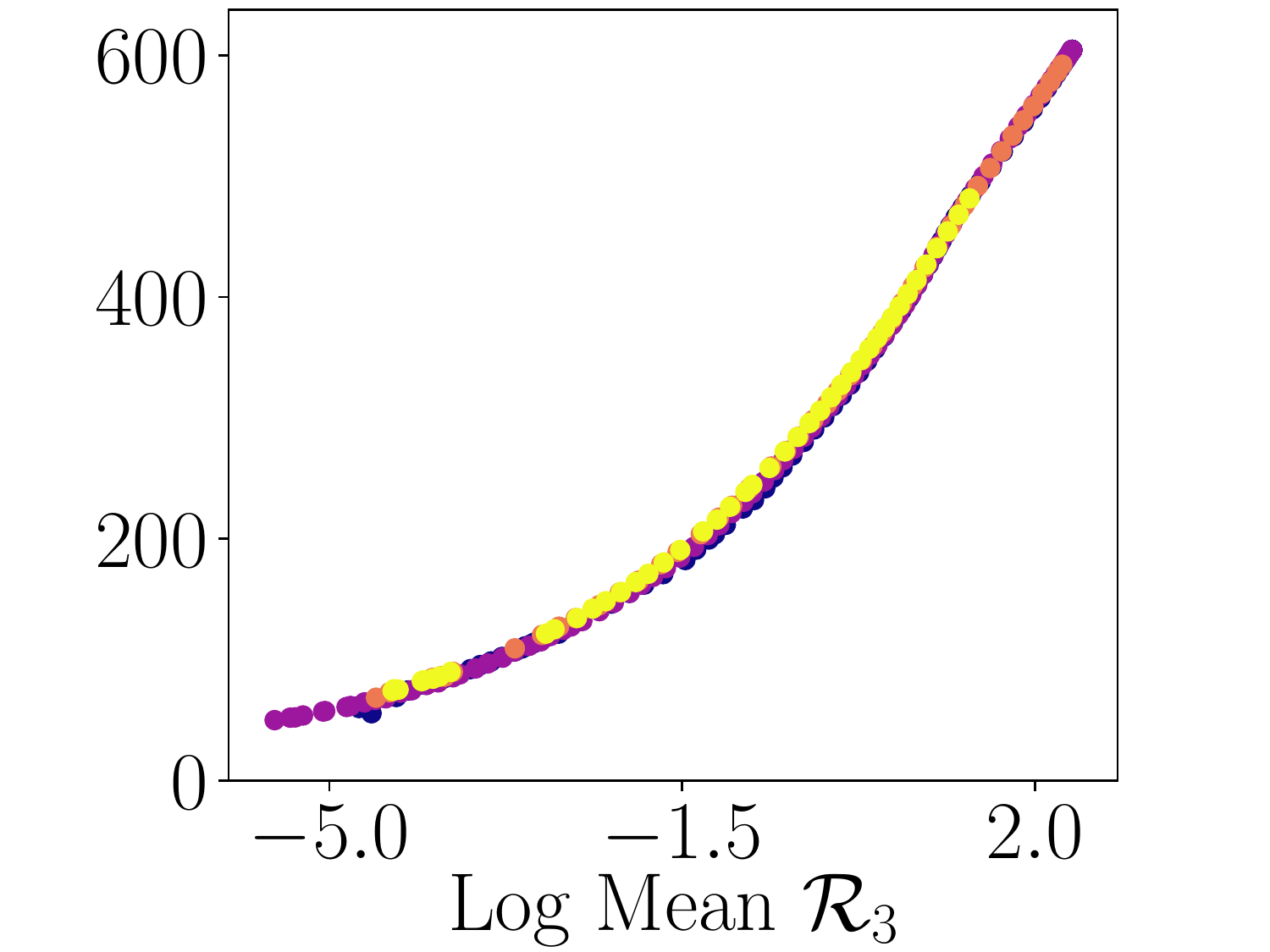}
        \caption{Order 3 Solver}
    \end{subfigure}
    \begin{subfigure}[b]{0.32\linewidth}
        \centering
        \includegraphics[width=1\linewidth]{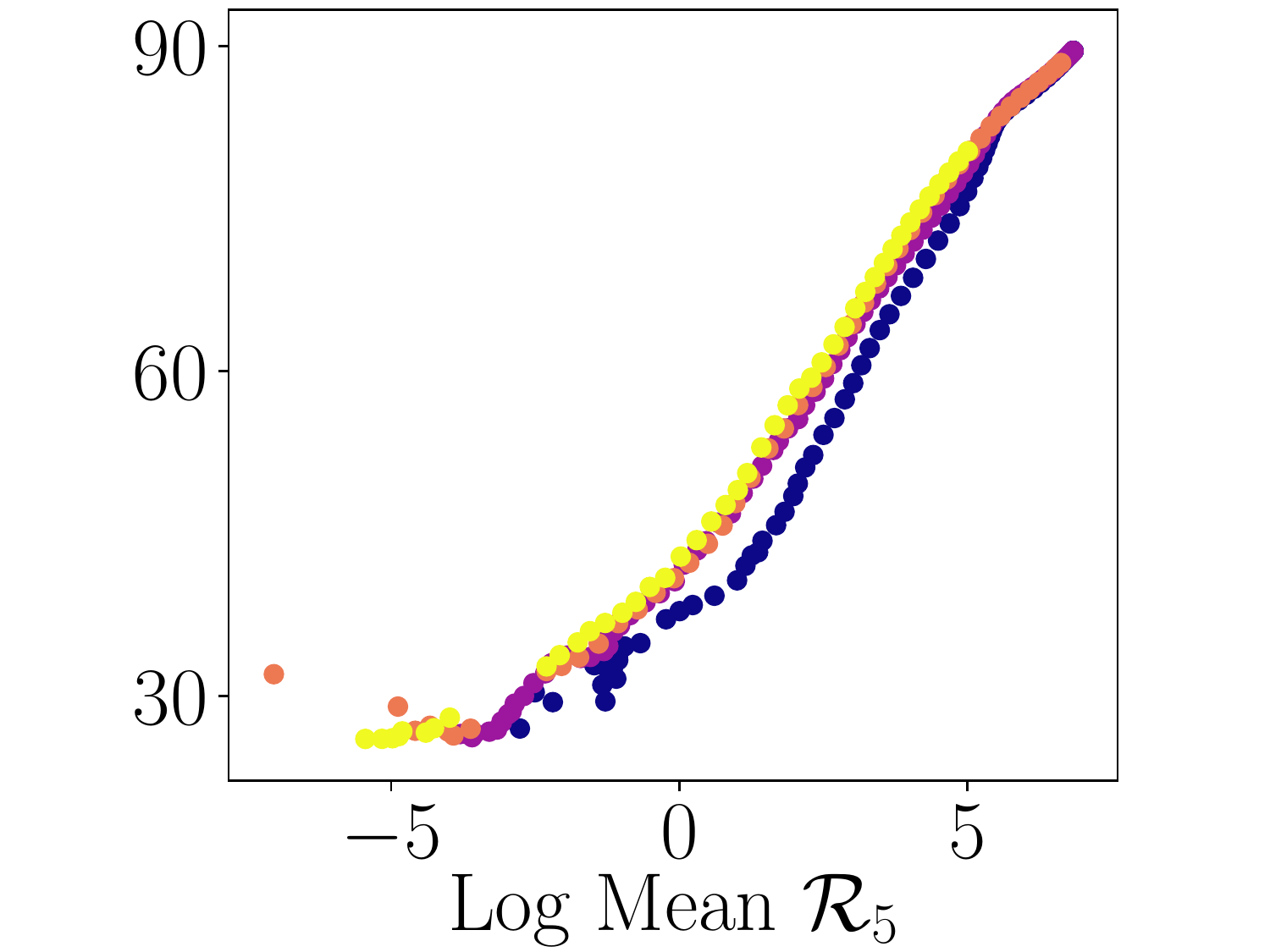}
        \caption{Order 5 Solver}
    \end{subfigure}
\caption{
For all orders, $\mathcal{R}_K$ varies monotonically with NFE.
For each order of solver, the model with the lowest NFE was achieved by regularizing the same order.
}\vspace{-1em}
\label{fig:regorders}
\end{figure}

Figures \ref{fig:orders} and \ref{fig:regorders} investigate this question on the task of MNIST classification. \Cref{fig:orders} compares the effectiveness of regularizing different orders when using a solver of a particular order.
For a 2nd order solver, regularizing $K = 2$ produces a strictly better trade-off between performance and speed, as expected.
For higher-order solvers, including ones with adaptive order, we found that regularizing orders above $K=3$ gave little benefit.

\Cref{fig:regorders} investigates the relationship between $\mathcal{R}_K$ and the quantity it is meant to be a surrogate for: NFE.
We observe a clear monotonic relationship between the two, for all orders of solver and regularization.

\subsection{Do we reduce training time?}
\label{finlay_comp}

Our approach produces models that are fastest to evaluate at test time. 
However, when we train with adaptive solvers we do not improve overall training time, due to the additional expense of computing our regularizer. 
Training with a fixed-grid solver is faster, but can be unstable if dynamics are unregularized.
\citet{finlay2020train}'s regularization and ours allow us to use fixed grid solvers and reduce training time. 
However, ours is 2.4$\times$ slower than \citet{finlay2020train} for FFJORD because their regularization re-uses terms already computed in the FFJORD training objective.
For objectives where these cannot be re-used, like MNIST classification, our method is $1.7\times$ slower, but achieves better test-time NFE.

\begin{table}[b]
    \vspace{-1.5em}
    \centering
    \caption{Density Estimation on MNIST using FFJORD.
    For adaptive solvers, indicated by $\infty$ Steps, our approach is slowest to train, but requires the fewest NFE once trained. 
    For fixed-step solvers our approach achieves lower bits/dim and NFE when comparing across fixed-grid solvers using the same number of steps.
    Fixed step solvers that diverged due to instability are indicated by \texttt{NaN} bits/dim. 
    }
    \begin{tabular}{lcccccccc}
         & \multicolumn{2}{c}{Training} & \multicolumn{5}{c}{Evaluation using adaptive solvers}  \\
         \cmidrule(lr){2-3}\cmidrule(lr){4-8} 
         & Steps & Hours & 
         Bits/Dim & \textsc{NFE} & $\mc{R}_2$ & 
         $\mathcal{B}$ & $\mathcal{K}$ \\
         \midrule
         Unregularized  
                            & 8  & -    & \verb|NaN|   & -      & -    & -    & -   \\
                            & $\infty$  & 35.8 & \bf{1.033} & 149  & 3596 & 4.76 & 73.6       \\
         \midrule
         RNODE 
                            & 5  & -    & \verb|NaN|   & -   & -     & -    & -  \\
        \citep{finlay2020train}   & 6  & \bf{8.4}  & 1.069 & 122 & 157.8 & 1.82 & 35.0\\
                            & 8  & 11.1 & 1.048 & 97  & 39.3  & 1.85 & 34.8 \\
                            & $\infty$  & 22.9 & 1.049 & 104  & 46.6  & 1.85 & 34.7 \\
          \midrule
         TayNODE (ours) 
                            & 5  & 20.3 & 1.077 & 98  & 31.3  & 2.89 & 36.5 \\
                            & 6  & 20.4 & 1.057 & 105 & 31.1  & 2.91 & 36.5 \\
                            & 8  & 27.1 & 1.046 & 98  & 26.0  & 2.53 & 36.3 \\
                            & $\infty$  & 54.7 & 1.039 & \bf{92}  & 22.9  & 2.41 & 36.2 \\
         \bottomrule
    \end{tabular}
    \label{table:ffjord_mnist}
\end{table}

\subsection{Are we making the solver overconfident?}
\begin{figure}[ht]
\centering
    \begin{subfigure}[c]{0.32\linewidth}
        \centering
        \includegraphics[width=1\linewidth, clip, trim=1cm 0.1cm 2cm 1.75cm]{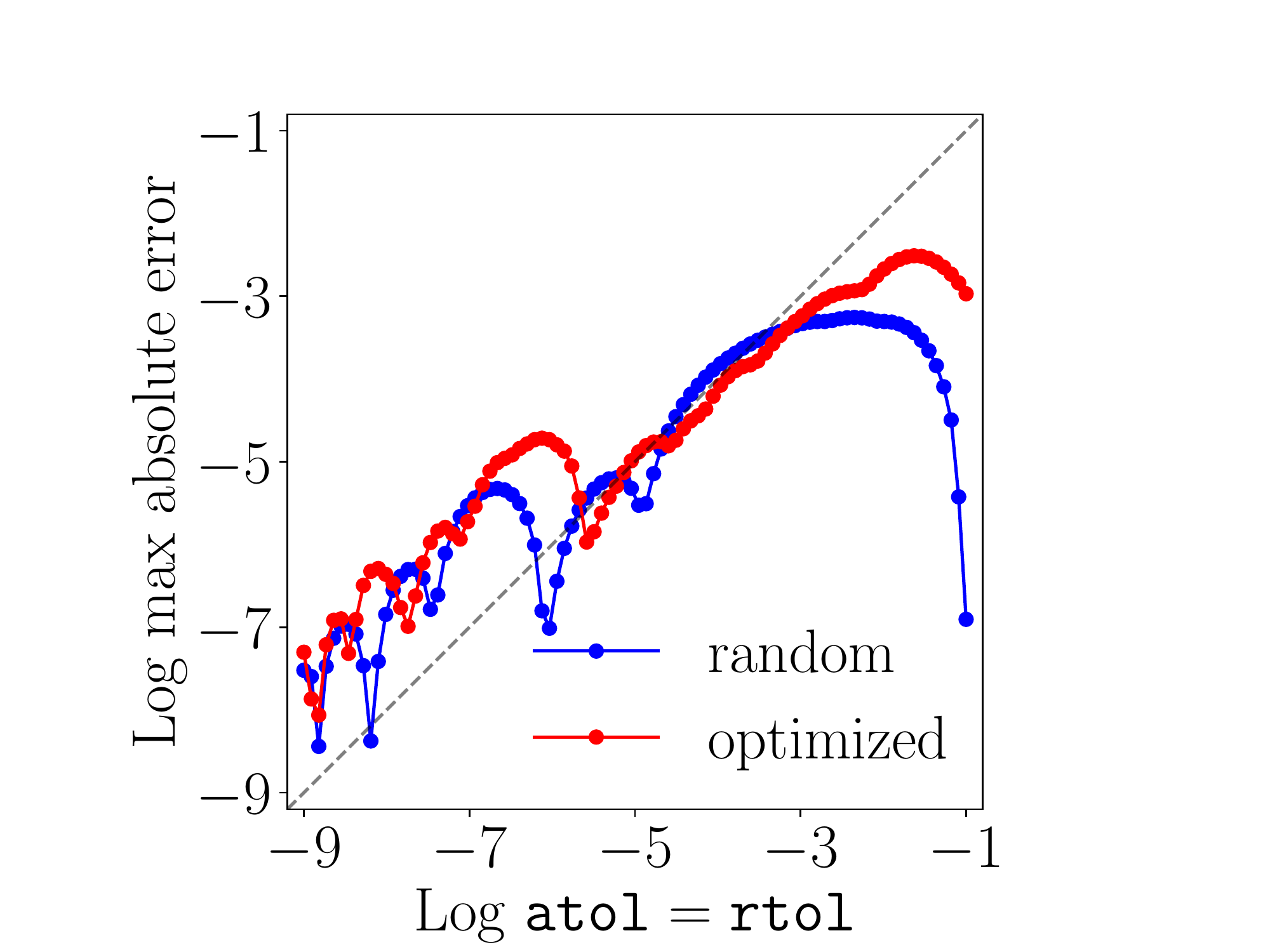}
        \caption{Solver tolerance vs. Error}
        \label{fig:attack}
    \end{subfigure}
    \begin{subfigure}[c]{0.32\linewidth}
        \centering
        \includegraphics[width=1\linewidth, clip, trim=0.1cm 0.1cm 0.2cm 0.3cm]{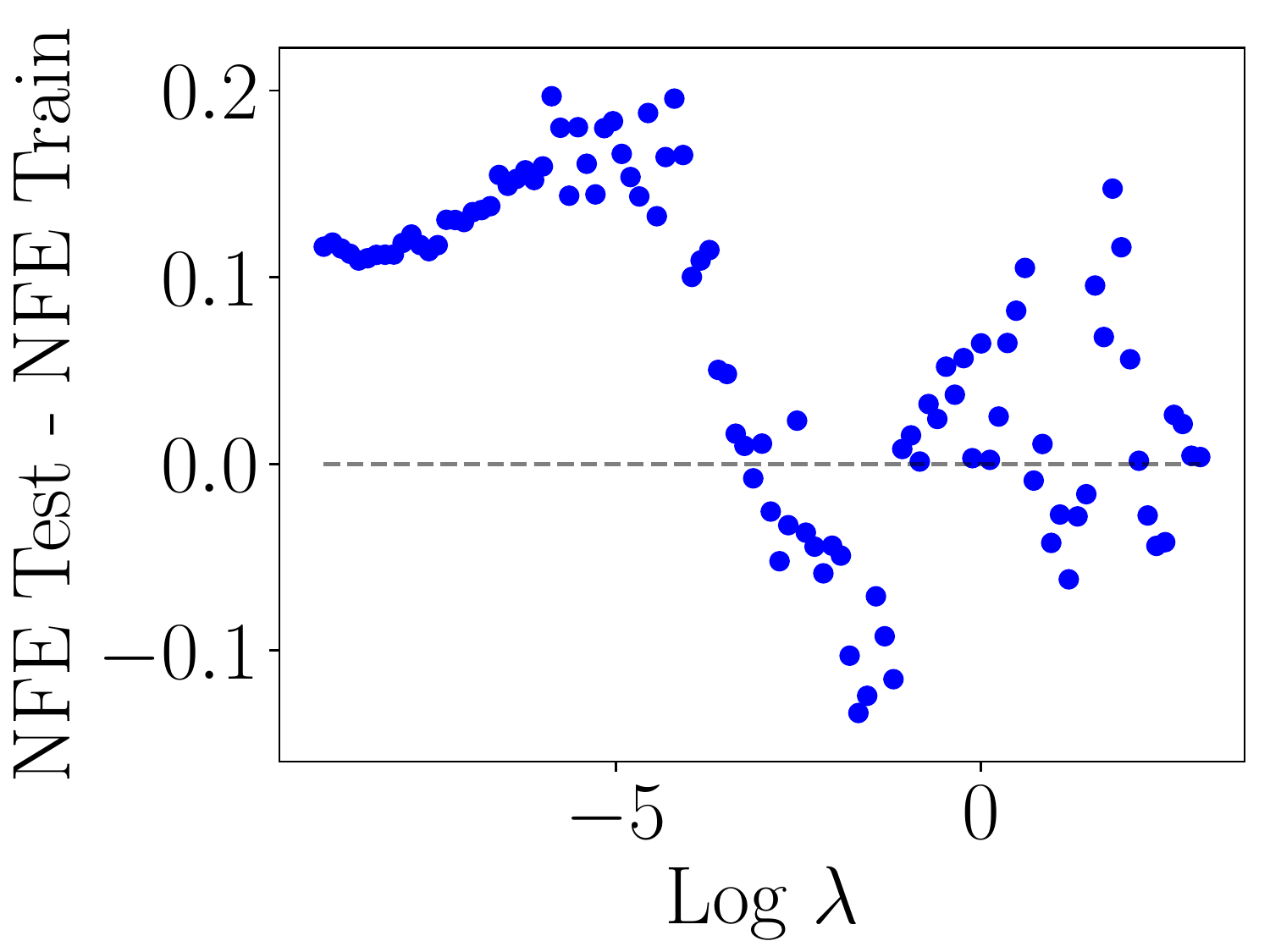}
        \caption{NFE Overfitting}
        \label{fig:overfitting}
    \end{subfigure}
    \begin{subfigure}[c]{0.32\linewidth}
        \centering
        \includegraphics[width=1\linewidth, clip, trim=0.4cm 0.4cm 2.4cm 1.8cm]{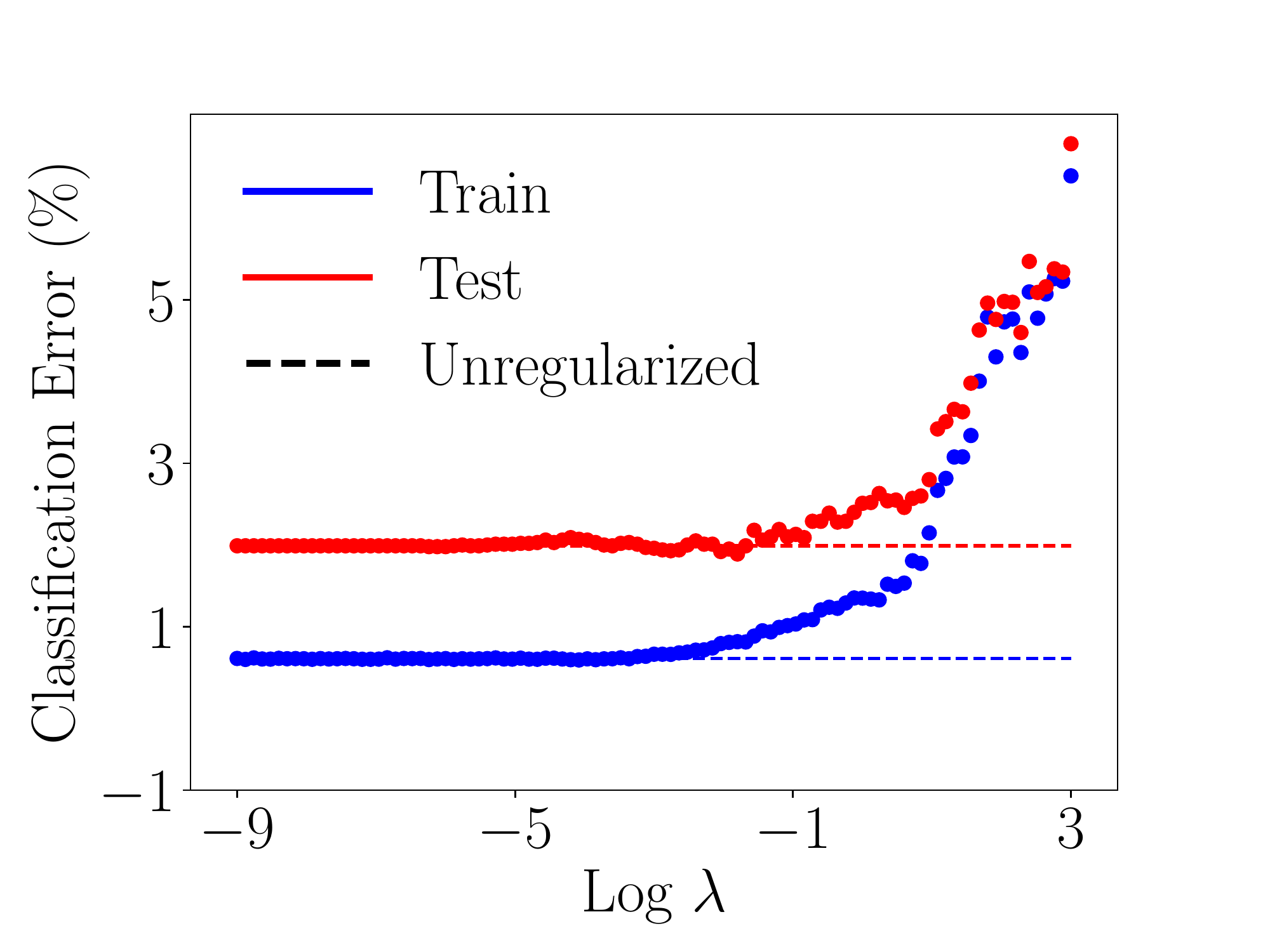}
        \caption{Statistical Regularization}
        \label{fig:stat_reg}
    \end{subfigure}
\captionsetup{belowskip=-15pt}
\caption{
\Cref{fig:attack}: We observe that the actual solver error is about equally well-calibrated for regularized dynamics as random dynamics, indicating that regularization does not make the solver overconfident.
\Cref{fig:overfitting}: There is negligible overfitting of solver speed.
\Cref{fig:stat_reg}: Speed regularization does not usefully improve generalization.
For large $\lambda$, our method reduces overfitting, but increases overall test error due to under-fitting.
}
\label{fig:misc}
\end{figure}

Because we optimize dynamics in a way specifically designed to make the solver take longer steps, we might fear that we are ``adversarially attacking'' our solver, making it overconfident in its ability to extrapolate.
\Cref{fig:attack} shows that this is not the case for MNIST classification.

\subsection{Does speed regularization overfit?}
\citet{finlay2020train} motivated one of their regularization terms by the possibility of overfitting: having faster dynamics only for the examples in the training set, but still low on the test set. 
However, they did not check whether overfitting was occurring.
In \cref{fig:overfitting} we confirm that our regularized dynamics have nearly identical average solve time on a held-out test set, on MNIST classification.

\section{Related Work}

\citet{grathwohl2019ffjord} mention attempting to use weight decay and spectral normalization to reduce NFE.
Of course, \citet{finlay2020train}, among other contributions, regularized trajectories of continuous normalizing flows and introduced the use of fixed-step solvers for stable and faster training.  
The use of fixed-step solvers is also explored in \citet{onken2020discretize}.
\citet{onken2020otflow} also regularized the trajectories of continuous normalizing flows, among other contributions.
\citet{massaroli2020dissecting} introduce new formulations of neural differential equations and investigate regularizing these models as applied to continuous normalizing flows in \citet{massaroli2020stableflows}.

\citet{poli2020hypersolvers} introduce solvers parameterized by neural networks for faster solving of neural differential equations. \citet{kidger2020hey} exploit the structure of the adjoint equations to construct a solver needing less NFE to speed up backpropagation through neural differential equations.

\citet{morrill2020neuralcde} introduce a method to improve the speed of neural controlled differential equations \citep{kidger2020neuralcde} which are designed for irregularly-sampled timeseries.

\citet{simard1991shaping} regularized the dynamics of discrete-time recurrent neural networks to improve their stability, by constraining the norm of the Jacobian of the dynamics function in the direction of its largest eigenvalue.
However, this approach has an $\mathcal{O}(D^3)$ time cost.
\citet{de2019gru} introduced a parameterization of neural ODEs analogous to instantaneous Gated Recurrent Unit (GRU) recurrent neural network architectures in order to stabilize training dynamics.
\citet{dupont2019augmented} provided theoretical arguments that adding extra dimensions to the state of a neural ODE should make training easier, and showed that this helped reduce NFE during training.

\citet{chang2017multi} noted the connection between residual networks and ODEs, and took advantage of this connection to gradually make resnets deeper during training, in order to save time.
One can view the increase in NFE with neural ODEs as an automatic, but uncontrolled, version of their method.
Their results suggest we might benefit from introducing a speed regularization schedule that gradually tapers off during training.

\citet{novak18, lecun92} regularized the gradients of neural networks to improve generalization.

We speculate on the application of our regularization in \cref{eq:reg_integral} for other purposes, including adversarial robustness \citep{yang2020interpolation, hanshu2019robustness}. and function approximation with Gaussian processes \citep{dutra2014maximum, van2008reproducing}. 


\section{Scope}
The initial speedups obtained in this paper are not yet enough to make neural ODEs competitive with standard fixed-depth architectures in terms of speed for standard supervised learning.
However, there are many applications where continuous-depth architectures provide a unique advantage.
Besides density models such as FFJORD and time series models, continuous-depth architectures have been applied in
solving mean-field games \citep{ruthotto19},
image segmentation \citep{pinckaers2019neural},
image super-resolution \citep{scao2020neural},
and molecular simulations \citep{wang2020differentiable}.
These applications, which already use continuous-time models, could benefit from the speed regularization proposed in this paper.

While we investigated only ODEs in this paper, this approach could presumably be extended straightforwardly to neural stochastic differential equations fit by adaptive solvers \citep{sdes2020} and other flavors of parametric differential equations fit by gradient descent \citep{rackauckas2019diffeqflux}.

\section{Limitations}


\paragraph{Hyperparameters}
The hyperparameter $\lambda$ needs to be chosen to balance speed and training loss.
One the other hand, neural ODEs don't require choosing the outer number of layers, which needs to be chosen separately for each stack of layers in standard architectures.

One also needs to choose solver order and tolerances, and these can substantially affect solver speed.
We did not investigate loosening tolerances, or modifying other parameters of the solver.
The default tolerance of \verb|1.4e-8| for both \verb|atol| and \verb|rtol| behaved well in all our experiments.

One also needs to choose $K$.
Higher $K$ seems to generally work better, but is slower per step at training time.
In principle, if one can express their utility explicitly in terms of training loss and NFE, it may be possible to tune $\lambda$ automatically during training using the predictable relationship between $\mathcal{R}_K$ and NFE shown in \cref{fig:regorders}.

\paragraph{Slower overall training}
Although speed regularization reduces the overall NFE during training, it makes each step more expensive.
In our density estimation experiments (\cref{table:ffjord_mnist}), the overall effect was about about 70\% slower training, compared to no regularization, when using adaptive solvers.
However, test-time evaluation is much faster, since there is no slowdown per step.


\section{Conclusions}
This paper is an initial attempt at controlling the integration time of differential equations by regularizing their dynamics.
This is an almost unexplored problem, and there are almost certainly better quantities to optimize than the ones examined in this paper.

Based on these initial experiments, we propose three practical takeaways:
\begin{enumerate}
    \item Across all tasks, tuning the regularization usually gave at least a 2x speedup without substantially hurting model performance.
    \item Overall training time with speed regularization is in general about 30\% to 50\% slower with adaptive solvers.
    \item For standard solvers, regularizing orders higher than $\mc{R}_2$ or $\mc{R}_3$ provided little additional benefit.
\end{enumerate}

\paragraph{Future work}
It may be possible to adapt solver architectures to take advantage of flexibility in choosing the dynamics.
Standard solver design has focused on robustly and accurately solving a given set of differential equations.
However, in a learning setting, we could consider simply rejecting some kinds of dynamics as being too difficult to solve, analogous to other kinds of constraints we put on models to encourage statistical regularization.


\section*{Acknowledgements}
We thank Dougal Maclaurin, Andreas Griewank, Barak Perlmutter, Ken Jackson, Chris Finlay, James Saunderson, James Bradbury, Ricky T.Q. Chen, Will Grathwohl, Chris Rackauckas, David Sanders, and Lyndon White for feedback and helpful discussions.
Resources used in preparing this research were provided, in part, by the Province of Ontario, the Government of Canada through CIFAR, NSERC, and companies sponsoring the Vector Institute.

\section*{Broader Impact}
We expect the main impact from this work, if any, would be through a potential improvement of the fundamental modeling tools of regression, classification, time series models, and density estimation. 
Thus the impact of this work is not distinct from that of improved machine learning tools in general. 
While machine learning tools present both benefits and unintended consequences, we avoid speculating further.

\bibliography{ref}
\bibliographystyle{icml2020}  

\newpage

\begin{appendices}

\section{Taylor-mode Automatic Differentiation}
\label{app:taylor}

\subsection{Taylor Polynomials}
\label{app:tay}

To clarify the relationship between the presentation in Chapter 13 of \citet{evalderivs} and our results we give the distinction between the Taylor coefficients and derivative coefficients, also known, unhelpfully, as \textit{Tensor} coefficients.

For a sufficiently smooth vector valued function $f : \mathbb{R}^n \rightarrow \mathbb{R}^m$ and the polynomial
\begin{equation}
\label{eqn:tayx-app}
x(t) = x_{[0]} + x_{[1]} t + x_{[2]} t^2 + x_{[3]} t^3 + \cdots + x_{[d]} t^d \in \mathbb{R}^n
\end{equation}
we are interested in the $d$-truncated Taylor expansion
\begin{align}
y(t) &= f(x(t)) + O(t^{d+1})\\
  \label{eqn:tayy-app}    &\equiv y_{[0]} + y_{[1]} t + y_{[2]} t^2 + y_{[3]} t^3 +\cdots + y_{[d]} t^d \in \mathbb{R}^m
\end{align}
with the notation that $y_{[i]} = \frac{1}{i!} y_i$ is the \textit{Taylor coefficient}, which is the normalized \textit{derivative coefficient} $y_i$.

The Taylor coefficients of the expansion, $y_{[j]}$, are smooth functions of the $i \leq j$ coefficients $x_{[i]}$, 
\begin{align}
y_{[0]} &= y_{[0]}(x_{[0]}) &&= f(x_{[0]})\\
y_{[1]} &= y_{[1]}(x_{[0]},x_{[1]})&& = f'(x_{[0]})x_{[1]}\\
y_{[2]} &= y_{[2]}(x_{[0]},x_{[1]},x_{[2]}) && = f'(x_{[0]}) x_{[2]} + \frac{1}{2} f''(x_{[0]}) x_{[1]} x_{[1]}\\
y_{[3]} &= y_{[3]}(x_{[0]},x_{[1]},x_{[2]},x_{[3]}) && = f'(x_{[0]}) x_{[3]} + f''(x_{[0]}) x_{[1]} x_{[2]} + \frac{1}{6} f'''(x_{[0]}) x_{[1]} x_{[1]} x_{[1]}\\
&&\vdots\nonumber
\end{align}

These, as given in \citet{evalderivs}, are written in terms of the normalized, Taylor coefficients. 
This obscures their direct relationship with the derivatives, which we make explicit.

Consider the polynomial \cref{eqn:tayx-app} with Taylor coefficients expanded so their normalization is clear. 
Further, let's use suggestive notation that these coefficients correspond to the higher derivatives of of $x$ with respect to $t$, making $x(t)$ a Taylor polynomial. 
That is $ x_{[i]} = \factfrac{i} x_i = \factfrac{i} \dnfrac{x}{t}{i}$.
\begin{align}
x(t) &= x_0 + x_1 t + \factfrac{2} x_2 t^2 + \factfrac{3} x_3 t^3 + \cdots + \factfrac{d} x_d t^d \in \mathbb{R}^n\\
     &= x_0 + \ddfrac{x}{t} t + \factfrac{2} \dnfrac{x}{t}{2} t^2 + \factfrac{3} \dnfrac{x}{t}{3} t^3 + \cdots + \factfrac{d} \dnfrac{x}{t}{d} t^d \in \mathbb{R}^n\\
\end{align}

Again, we are interested in the polynomial \cref{eqn:tayy-app}, but with the normalization terms explicit
\begin{equation}
  y(t) \equiv y_0 + y_1 t + \frac{1}{2!} y_2 t^2 + \frac{1}{3!} y_3 t^3 +\cdots + \factfrac{d} y_d t^d \in \mathbb{R}^m \label{eqn:derexp}
\end{equation}

Now we can expand the expressions for the Taylor coefficients $y_{[i]}$ to expressions for derivative coefficients $y_i = i! y_\tayc{i}$

The coefficients of the Taylor expansion, $y_j$, are smooth functions of the $i \leq j$ coefficients $x_i$, 
\begin{align}
y_0 &= y_0(x_0) &&= y_\tayc{0}(x_0) \nonumber\\
\label{eqn:d0}& &&=  f(x_0)\\
y_1 &= y_1(x_0,x_1)&& = y_\tayc{1}(x_0,x_1) \nonumber\\
& && = f'(x_0)x_1 \nonumber \\ 
\label{eqn:d1}& && =f'(x_0) \ddfrac{x}{t}\\
y_2 &= y_2(x_0,x_1,x_2) && = 2! \left( y_\tayc{2}(x_0,x_1,\factfrac{2} x_2)\right) \nonumber \\ 
& &&= 2! \left(f'(x_0) \factfrac{2} x_2 + \frac{1}{2} f''(x_0) x_1 x_1\right)\nonumber\\
& &&= f'(x_0)x_2 + f''(x_0) x_1 x_1\nonumber\\
\label{eqn:fdbd2}& && = f'(x_0) \dnfrac{x}{t}{2} +  f''(x_0) \left(\ddfrac{x}{t}\right)^2\\
\label{eqn:d2}& && = \dnfrac{}{t}{2}f(x(t))\\
y_3 &= y_3(x_0,x_1,x_2,x_3)&&= 3!\left( y_\tayc{3}(x_0,x_1,\factfrac{2}x_2,\factfrac{3}x_3)\right)\nonumber\\
& && = 3! \left( f'(x_0) \factfrac{3}x_3 + f''(x_0) x_1 \factfrac{2} x_2 + \frac{1}{6} f'''(x_0) x_1 x_1 x_1 \right)\nonumber\\
& && = f'(x_0)x_3 + 3 f''(x_0) x_1 x_2 + f'''(x_0) x_1 x_1 x_1\nonumber\\
\label{eqn:fdbd3}& && = f'(x_0)\dnfrac{x}{t}{3} + 3 f''(x_0) \ddfrac{x}{t}\dnfrac{x}{t}{2} + f'''(x_0)\left(\ddfrac{x}{t}\right)^3\\
\label{eqn:d3}& && = \dnfrac{}{t}{3}f(x(t))\\
&&\vdots \nonumber
\end{align}

Therefore, \cref{eqn:d0,eqn:d1,eqn:d2,eqn:d3} show that the derivative coefficient $y_i$ are exactly the $i$th order higher derivatives of the composition $f(x(t))$ with respect to $t$.
The key insight to this exercise is that by writing the derivative coefficients explicitly we reveal that the expressions for the terms, \cref{eqn:d0,eqn:d1,eqn:fdbd2,eqn:fdbd3}, involve terms previously computed for lower order terms.

In general, it will be useful to consider that the $y_k$ derivative coefficients is a function of all lower order input derivatives 
\begin{equation}
  y_k = y_k(x_0, \dots, x_k).
  \label{eq:coefffunc}
\end{equation}
We provide the API to compute this in JAX by indexing the $k$-output of \jettt 
$$ y_k = \jetcode{f}{x_0}{x_1,\dots,x_k}[k].$$

\subsection{Relationship with Differential Equations}
\subsubsection{Autonomous Form}
\label{app:autoform}

We can transform the initial value problem
\begin{equation}
  \ddfrac{x}{t} = f(x(t),t) \quad \text{where} \quad x(t_0) = x_0 \in \R{n}
\end{equation}
into an \textit{autonomous} dynamical system by augmenting the system to include the independent variable
with trivial dynamics \cite{hairer}:
\begin{equation}
  \ddfrac{}{t}\begin{pmatrix}x\\ t\end{pmatrix} = \begin{pmatrix}f(x(t))\\ 1 \end{pmatrix} \quad \text{where} \quad \begin{pmatrix}x(0)\\ t(0)\end{pmatrix} = \begin{pmatrix}x_0 \\ t_0\end{pmatrix}\in \R{n}
\end{equation}

We do this for notational convenience, as well it disambiguates that derivatives with respect to $t$ are meant in the ``total" sense.
This is aleviates the potential ambiguity of $\pfrac{}{t} f(x(t),t)$ which could
mean both the derivative with respect to the second argument
and the derivative through $x(t)$ by the chain rule $\pfrac{f}{x}\pfrac{x}{t}$.

\subsubsection{Taylor Coefficients for ODE Solution with \jettt}
\label{app:recsol}

Recall that \jettt\ gives us the coefficients for $y_i$ as a function of $f$ and the coefficients $x_{j\leq i}$. 
We can use \jettt\  and the relationship $x_{k+1} = y_k$ to recursively compute the coefficients of the solution polynomial. 

\begin{algorithm}
\caption{Taylor Coefficients for ODE Solution by Recursive Jet}
\label{alg:recsoljet}
\begin{minted}{python}
# Have: x_0, f
# Want: x_1, ..., x_K

y_0 = jet(f, x_0, (0,)) # equivalently, f(x_0)
x_1 = y_0

for k in range(K):
  (y_0, (y_1,..., y_k)) = jet(f, x_0, (x_1,..., x_k))
  x_{k+1} = y_k

return x_0, (x_1, ..., x_K)
\end{minted}
\end{algorithm}

\subsection{Regularizing Taylor Terms}

Computing the Taylor coefficients for the ODE solution as in \cref{alg:recsoljet} will give a local approximation to the ODE solution.
If infinitely many Taylor coefficients could be computed this would give the exact solution.
The order of the final Taylor coefficient, determining the truncation of the polynomial, gives the order of the approximation.

\begin{figure}[h]
  \centering
  \includegraphics[width=0.8\linewidth]{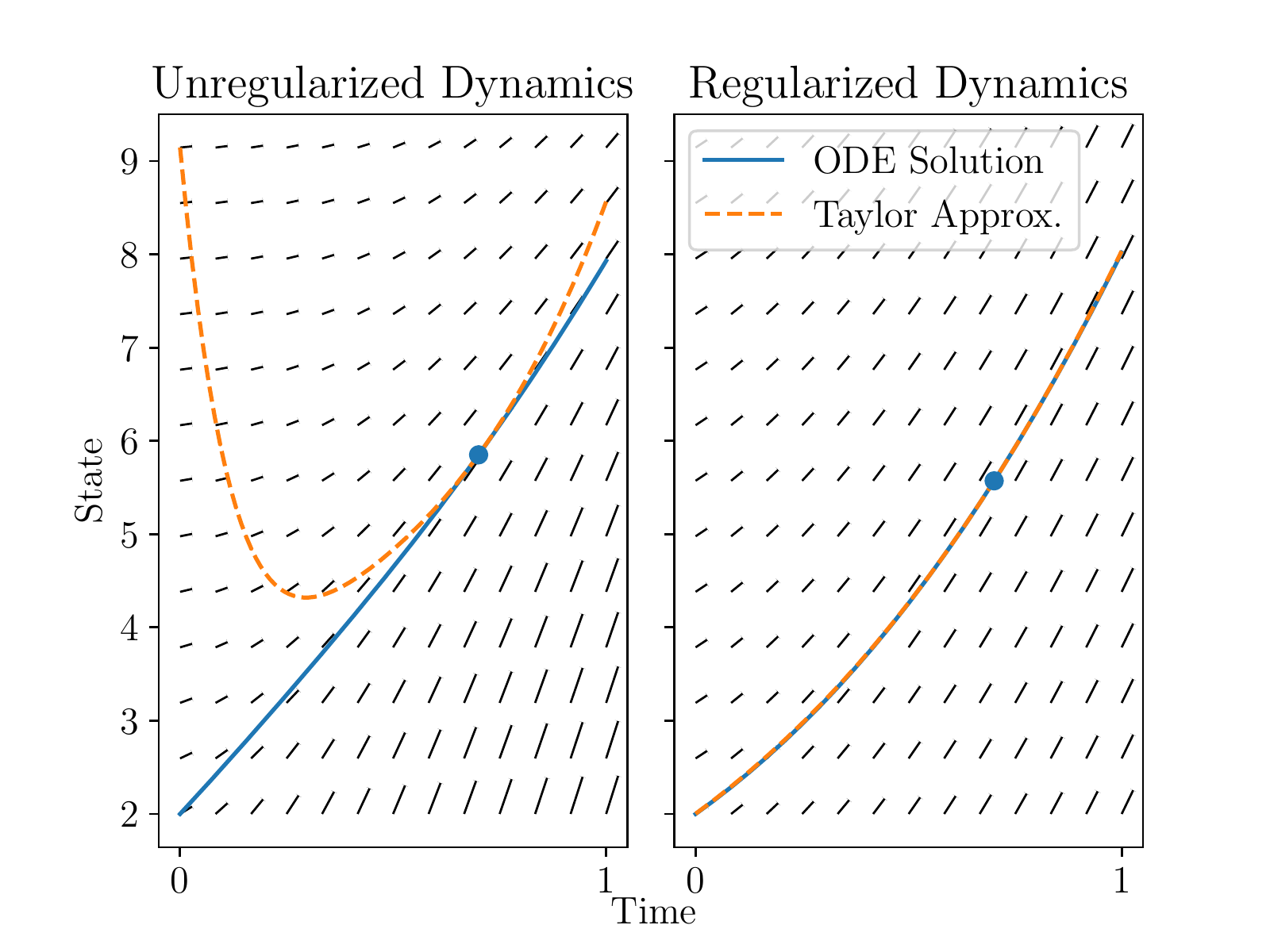}
  \caption{
\emph{Left:} The dynamics and a trajectory of a neural ODE trained on a toy supervised learning problem.  The dynamics are poorly approximated by a 6th-order local Taylor series, and requires 92 NFE by a solve by a 5th-order Runge-Kutta solver.  \emph{Right:}
Regularizing the 6th-order derivatives of trajectories gives dynamics that are easier to solve numerically, requiring only 68 NFE.
}
\label{fig:tay_mlp}
\end{figure}

If the higher order Taylor coefficients of the solution are large, then truncation will result in a local approximation that quickly diverts from the solution.
However, if the higher Taylor coefficients are small then the local approximation will remain close to the solution.
This motivates our regularization method. The effect of our regularizer on the Taylor expansion of a solution to a neural ODE can be seen in \cref{fig:tay_mlp}.

\section{Experimental Details}
\label{appendix_experiment}

Experiments were conducted using GPU-based ODE solvers.
Training gradients were computed using the adjoint method, in which the trajectory is reconstructed backwards in time to save memory, for backpropagation.
As in \citet{finlay2020train}, we normalize our regularization term in \cref{eq:reg_integral} by the dimension of the vector-valued trajectory $\mb{z}(t)$ so that we may choose $\lambda$ free of scaling by the dimension of the problem.

\subsection{Efficient computation of the gradient of regularization term}

To optimize our regularized objective, we must compute its gradient. We use the adjoint method as described in \cite{chen2018neural} to differentiate through the solution to the ODE. In particular, to optimize our model we only need to compute the gradient of the regularization term. The adjoint method gives the gradient of the ODE solution as a solution to an augmented ODE.

\subsection{Supervised Learning}
\label{appendix_experiment_mnist}

The dynamics function $f: \R{d}\times\R{} \to \R{d}$ is given by an MLP as follows

\begin{gather*}
    z_1 = \sigma(x) \\
    h_1 = W_1[z_1 ; t] + b_1 \\
    z_2 = \sigma(h_1) \\
    y = W_2[z_2 ; t] + b_2
\end{gather*}

Where $[\cdot;\cdot]$ denotes concatenation of a scalar onto a column vector. The parameters are $W_1 \in \R{h\times d}, b_1 \in \R{h}$ and $W_2 \in \R{d\times h}, b_2 \in \R{d}$. Here we use 100 hidden units, i.e. $h=100$. We have $d=784$, the dimension of an MNIST image.

We train with a batch size of 100 for 160 epochs. We use the standard training set of 60,000 images, and the standard test set of 10,000 images as a validation/test set. We optimize our model using SGD with momentum with $\beta=0.9$. Our learning rate schedule is \verb|1e-1| for the first 60 epochs, \verb|1e-2| until epoch 100, \verb|1e-3| until epoch 140, and \verb|1e-4| for the final 20 epochs.

\subsection{Continuous Generative Modelling of Time-Series}

\label{appendix_experiment_latent}

The PhysioNet dataset consists of observations of 41 distinct traits over a time period of 48 hours. We remove the parameters \enquote{Age}, \enquote{Gender}, \enquote{Height}, and \enquote{ICUType} as these attributes do not vary in time. We also quantize the measurements for each attribute by the hour by averaging multiple measurements within the same hour. This leaves 49 unique time stamps (the extra time stamp for observations at exactly the endpoint of the 48 hour observation period). We report all our losses on this quantized data. We performed this rather coarse quantization for computational reasons having to do with our particular implementation of this model. The validation split was obtained by taking a random split of 20\% of the trajectories from the full dataset. In total there are 8000 trajectories. Code is included for processing the dataset, and links to downloading the data may be found in the code for \citet{rubanova2019latent}. All other experimental details may be found in the main body and appendices of \citet{rubanova2019latent}.

\subsection{Continuous Normalizing Flows}

For the model trained on the MINIBOONE tabular dataset from \citet{tabularpapamakarios}, we used the same architecture as in Table 4 in the appendix of \citet{grathwohl2019ffjord}. We chose the number of epochs and a learning rate schedule based on manual tuning on the validation set, in contrast to \citet{grathwohl2019ffjord} who tuned these automatically using early stopping and an automatic heuristic for the learning rate decay using evaluation on a validation set. In particular, we trained for 500 epochs with a learning rate of \verb|1e-3| for the first 300 epochs, \verb|1e-4| until epoch 425, and \verb|1e-5| for the remaining 75 epochs. The number of epochs and learning rate schedule was determined by evaluating the model on the validation set every 10 epochs, and decaying the learning rate by a factor of 10 once the loss on the validation set stopped improving for several evaluations, with the goal of matching or improving upon the log-likelihood reported in \citet{grathwohl2019ffjord}. The data was obtained as made available from \citet{tabularpapamakarios}, which was already processed and split into train/validation/test. In particular, the training set has 29556 examples, the validation set has 3284 examples, and the test set has 3648 examples, which consist of 43 features.

It is important to note that we implemented a single-flow model for the MNIST dataset, while the original comparison in \citet{finlay2020train} was on a multi-flow model. This accounts for discrepancy in bits/dim and NFE reported in \citet{finlay2020train}.

All other experimental details are as in \citet{grathwohl2019ffjord}. 

\subsection{Hardware}

MNIST Supervised learning, Physionet Time-series, and MNIST FFJORD experiments were trained and evaluated on NVIDIA Tesla P100 GPU. Tabular data FFJORD experiments were evaluated on NVIDIA Tesla P100 GPU but trained on NVIDIA Tesla T4 GPU. All experiments except for MNIST FFJORD were trained with double precision for purposes of reproducibility.


\section{Additional Results}
\label{app:more experiments}

\subsection{Overfitting of NFE}
    
\begin{figure}[h]
  \centering
  \includegraphics[width=0.4\linewidth]{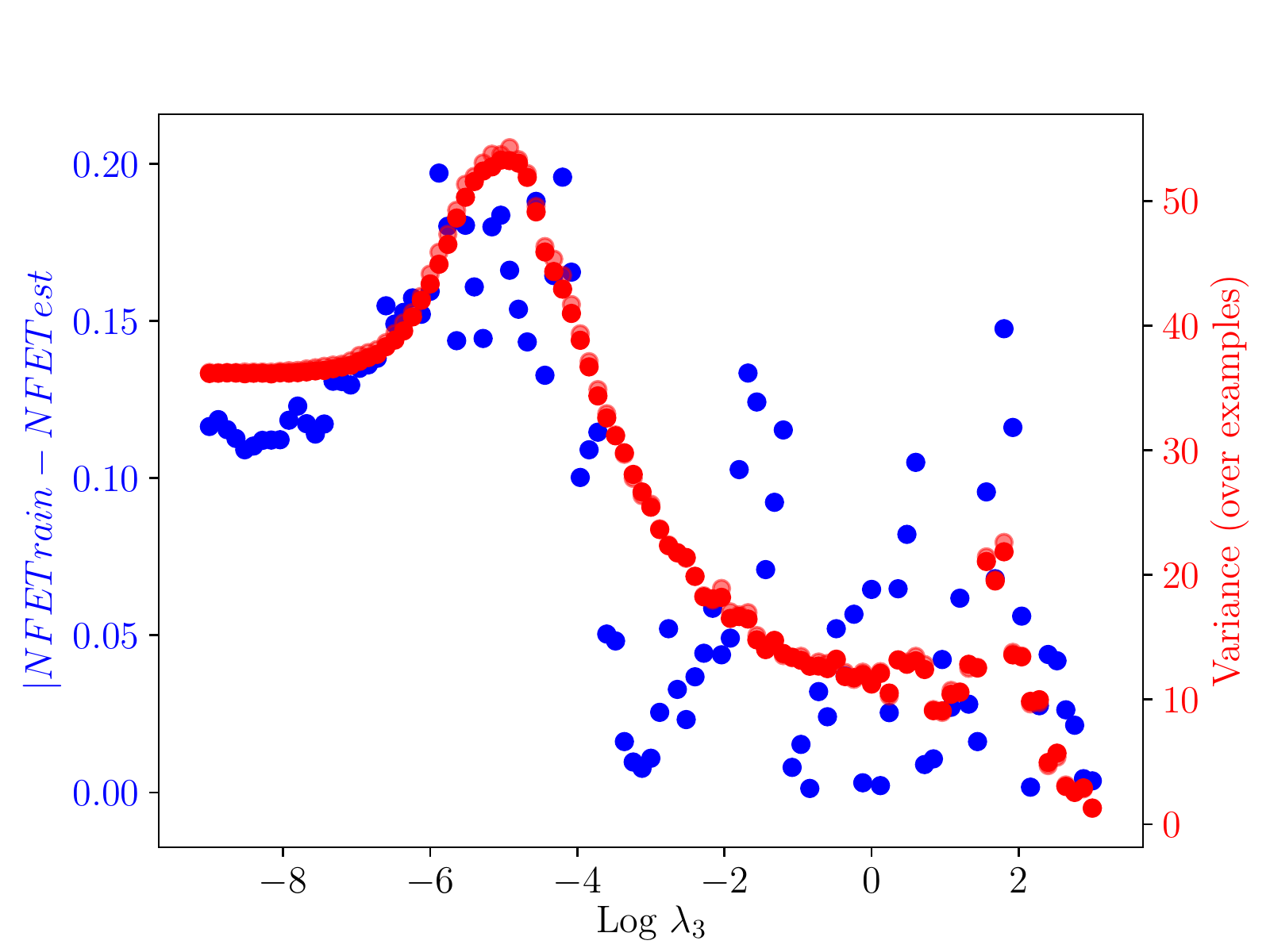}
\caption{The difference in NFE is tracked by the variance of NFE.}
\label{fig:appendix_pareto_test1}
\end{figure}

In \cref{fig:appendix_pareto_test1} we note that there is a striking correspondence in the variance of NFE across individual examples (in both the train set (dark red) and test set (light red)) and the absolute difference in NFE between examples in the training set and test set. This suggests that any difference in the average NFE between training examples and test examples is explained by noise in the estimate of the true average NFE. It is also interesting that speed regularization does not have a monotonic relationship with the variance of NFE, and we speculate as to how this might interact between the correspondence of NFE for a particular example and the difficulty in the model correctly classifying it.

\subsection{Trading off function evaluations with a surrogate loss}
\begin{figure}[!htb]
            \minipage{0.49\textwidth}
              \includegraphics[width=\linewidth]{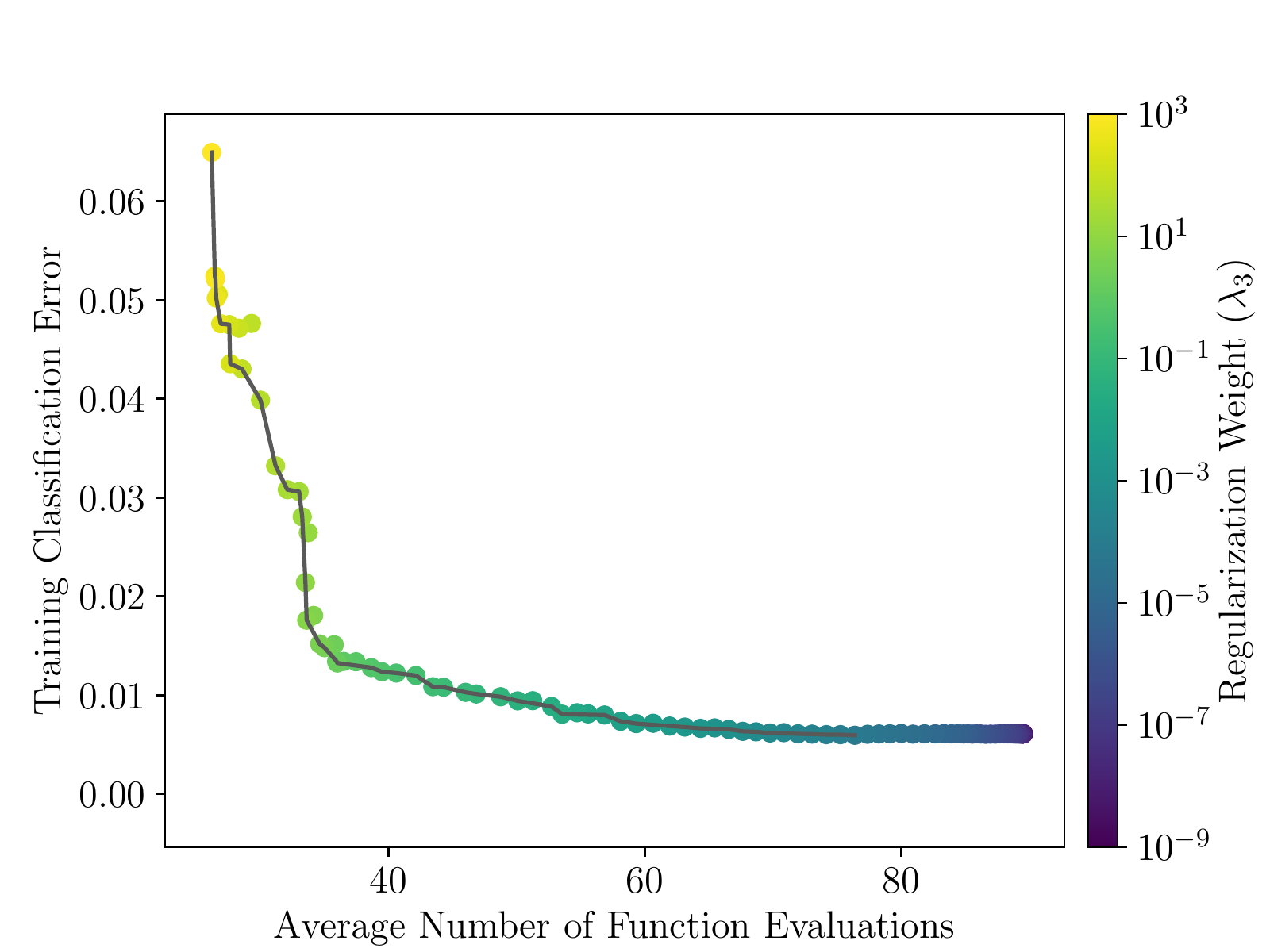}
              \caption{MNIST Classification}\label{fig:appendix_awesome_image1}
            \endminipage\hfill
            \minipage{0.49\textwidth}
              \includegraphics[width=\linewidth]{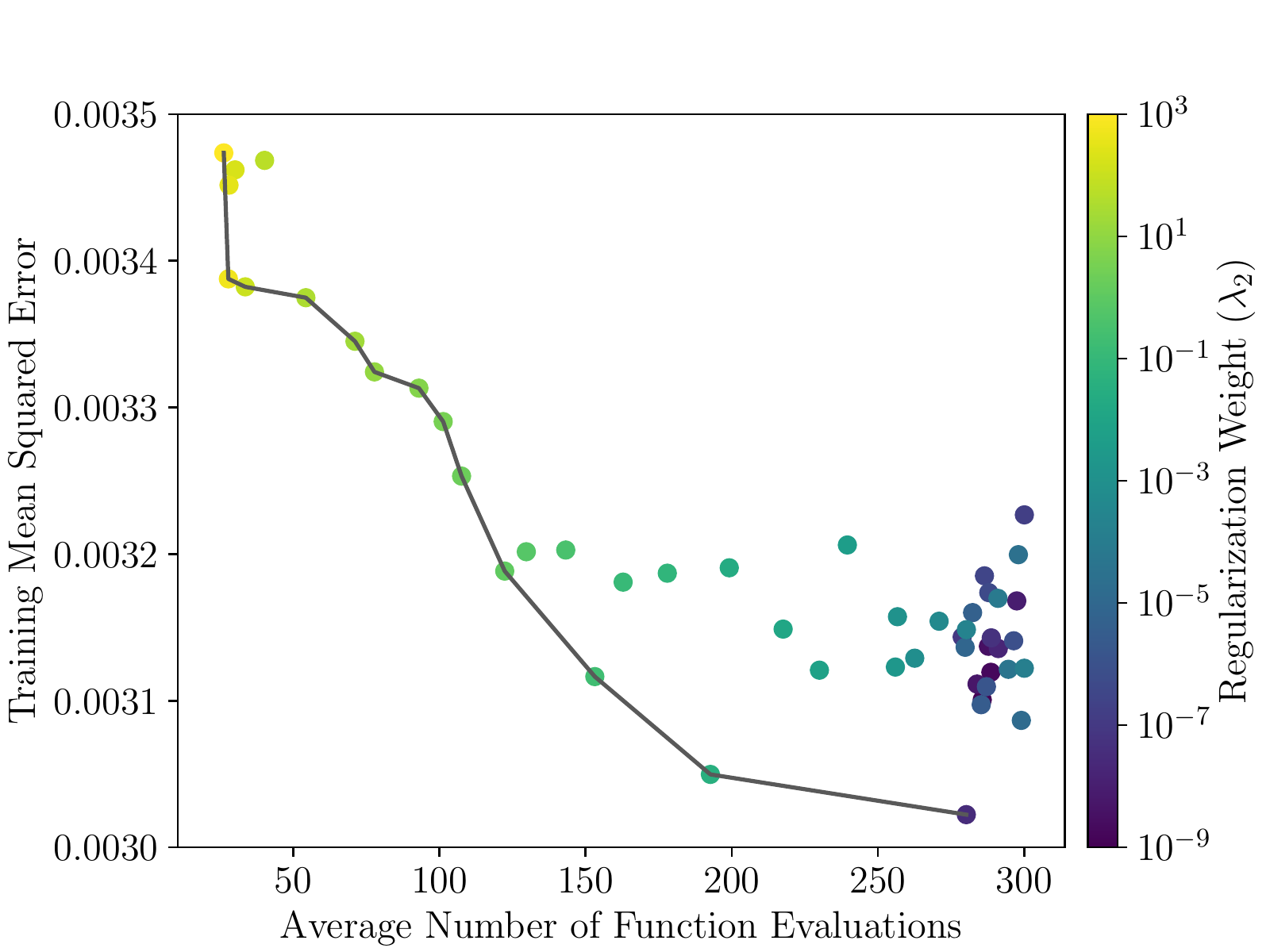}
              \caption{Physionet Time-Series}\label{fig:appendix_awesome_image2}
            \endminipage
            \end{figure}

            In \cref{fig:appendix_awesome_image1} and \cref{fig:appendix_awesome_image2} we confirm that our method poses a suitable tradeoff not only on the loss being optimized, but also on the potentially non-differentiable loss which we truly care about. On MNIST, we get a similar pareto curve when plotting classification error as opposed to cross-entropy loss, and similarly on the time-series modelling task we see that we get a similar pareto curve on MSE loss as compared to IWAE loss. The pareto curves are plotted for $\mc{R}_3$, $\mc{R}_2$ respectively.
            
\subsection{Wall-clock Time}

We include additional tables with wall-clock time and training with fixed grid solvers in \cref{mnist_class_table} and \cref{tabular_table}.

    \begin{table}[b]
        \vspace{-1.5em}
        \centering
        \caption{Classification on MNIST}
        \label{mnist_class_table}
        \begin{tabular}{lcccccccc}
             & \multicolumn{2}{c}{Training} & \multicolumn{5}{c}{Evaluation using adaptive solvers}  \\
             \cmidrule(lr){2-3}\cmidrule(lr){4-8} \\
             & Steps & Hours & 
             Loss & NFE & $\mc{R}_2$ & $\mc{B}$ & $\mc{K}$ \\
             \midrule
             No Regularization  
                                & 2  & 0.08 & .0239 & 116 & 25.9 & .231 & 7.91\\
                                & 4  & 0.13 & .0235 & 110 & 21.9 & .234 & 7.66\\
                                & 8  & 0.23 & .0236 & 110 & 21.3 & .233 & 7.62 \\
                                & $\infty$  & 1.71 & .0235 & 110 & -    & .233 & 7.63\\
             \midrule
            RNODE              & 2  & 0.12 & .0238 & 110 & 18.4 & .229 & 7.07\\
             \citep{finlay2020train} 
                                & 4  & 0.20 & .0238 & 110 & 14.6 & .230 & 6.85\\
                                & 8  & 0.37 & .0238 & 110 & 14.1 & .229 & 6.82 \\
             \midrule
             TayNODE (ours)     
                                & 2  & 0.19 & .0234 & 104 & 3.2  & .217 & 7.12\\
                                & 4  & 0.33 & .0234 & 104 & 2.4  & .218 & 7.06\\
                                & 8  & 0.61 & .0234 & 104 & 2.4  & .219 & 7.06 \\
                                & $\infty$  & 2.56 & .0234 & 104 & -    & .233 & 7.63\\
             \bottomrule
        \end{tabular}
    \end{table}
    
    \begin{table}
        \vspace{-1.5em}
        \centering
        \caption{Density Estimation on Tabular Data (MINIBOONE)}
        \label{tabular_table}
        \begin{tabular}{lcccccccc}
             & \multicolumn{2}{c}{Training} & \multicolumn{5}{c}{Evaluation using adaptive solvers}  \\
             \cmidrule(lr){2-3}\cmidrule(lr){4-8} \\
             & Steps & Hours & 
             Loss & NFE & $\mc{R}_2$ & $\mc{B}$ & $\mc{K}$ \\
             \midrule
             No Regularization  
                                & 4   & 0.19 & 9.78  & 185 & 17.1 & 4.10 & 1.72\\
                                & 8   & 0.37 & 9.77  & 184 & 19.0 & 4.10 & 1.77\\
                                & $\infty$   & 1.64 & 9.74  & 182 & -    & 4.10 & 1.77\\
             \midrule
             RNODE 
                                & 4   & 0.19 & 9.77 & 182 & 15.9 & 4.02 & 1.65\\
             \citep{finlay2020train}
                                & 8   & 0.38 & 9.76 & 181 & 17.3 & 4.01 & 1.69\\
                                & 16  & 0.73 & 9.77 & 189 & 17.5 & 4.03 & 1.70\\
             \midrule
            TayNODE (ours)     
                                & 4   & 0.49 & 9.84 & 177 & 13.1 & 4.00 & 1.57\\
                                & 8   & 0.96 & 9.79 & 181 & 13.6 & 3.99 & 1.58\\
                                & 16  & 1.90 & 9.77 & 181 & 13.7 & 3.99 & 1.59\\
             \bottomrule
        \end{tabular}
    \end{table}

        
        
        
        
        

    
    
    \section{Comparison to How to Train Your Neural ODE}

    The terms from \citet{finlay2020train} are
    
    \begin{align*}
        \left\Vert f(\mb{z}(t), t, \theta) \right\Vert^2_2 
    \end{align*}
    
    and an estimate of
    
    \begin{align*}
        \left\Vert \nabla_{\mb{z}}f(\mb{z}(t), t, \theta) \right\Vert^2_F 
    \end{align*}
    
    These are combined with a weighted average and integrated along the solution trajectory.
    
    These terms are motivated by the expansion  
    
    \begin{align*}
        \Ddn{\mb{z}(t)}{t}{2} & = \nabla_\mb{z} f(\mb{z}(t), t)f(\mb{z}(t), t) + \dd{f}{t}(\mb{z}(t), t)
    \end{align*}
    
    Namely, \cref{fin_kin} regularizes the first total derivative of the solution, $f(\mb{z}(t), t)$, along the trajectory, and \cref{fin_fro} regularizes a stochastic estimate of the Frobenius norm of the spatial derivative, $\nabla_\mb{z} f(\mb{z}(t), t)$, along the solution trajectory.
    
    In contrast, $\mc{R}_2$ regularizes the norm of the second total derivative directly. In particular, this takes into account the $\dd{f}{t}$ term. In other words, this accounts for the explicit dependence of $f$ on time, while \cref{fin_kin} and \cref{fin_fro} capture only the implicit dependence on time through $\mb{z}(t)$. 
    
    Even in the case of an autonomous system, that is, where $\dd{f}{t}$ is identically 0 and the dynamics $f$ only depend implicitly on time, these terms still differ. Namely, $\mc{R}_2$ integrates the following along the solution trajectory:
    
    \begin{align*}
        \left\Vert \nabla_{\mb{z}}f(\mb{z}(t), t, \theta)f(\mb{z}(t), t, \theta) \right\Vert^2_2 
    \end{align*}
    
    while \citet{finlay2020train} penalizes the respective norms of the matrix $\nabla_\mb{z}f(\mb{z}(t), t)$ and vector $f(\mb{z}(t), t)$ separately.
    
    


\end{appendices}

\end{document}